\documentclass{article}
\usepackage{ijcai26}

\usepackage{times}
\usepackage{soul}
\usepackage{url}
\usepackage[hidelinks]{hyperref}
\usepackage[utf8]{inputenc}
\usepackage[small]{caption}
\usepackage{graphicx}
\usepackage{amsmath}
\usepackage{amsthm}
\usepackage{booktabs}
\usepackage{algorithm}
\usepackage{algorithmic}
\usepackage[switch]{lineno}

\usepackage{subfigure}
\usepackage{booktabs}
\usepackage{amssymb}

\theoremstyle{plain}
\newtheorem{theorem}{Theorem}[section]
\newtheorem{proposition}[theorem]{Proposition}
\newtheorem{lemma}[theorem]{Lemma}
\newtheorem{corollary}[theorem]{Corollary}
\theoremstyle{definition}
\newtheorem{definition}[theorem]{Definition}

\theoremstyle{remark}

\usepackage{makecell}

\title{Aligning the Spectrum: Hybrid Graph Pre-training and Prompt Tuning across Homophily and Heterophily}

\author{
    Haitong Luo$^{1,2}$ \and
    Suhang Wang$^3$ \and
    Weiyao Zhang$^1$ \and
    Ruiqi Meng$^{1, 2}$ \and
    Xuying Meng$^{1\ast}$ \and
    Yujun Zhang$^1$\thanks{Corresponding Authors} \and
\affiliations 
    $^1$Institute of Computing Technology, Chinese Academy of Sciences,\\
    $^2$University of Chinese Academy of Sciences,\\
    $^3$Pennsylvania State University\\
\emails
    \{luohaitong21s, mengxuying, nrcyujun\}@ict.ac.cn
}

\begin{document}

\maketitle

\begin{abstract}

Graph ``pre-training and prompt-tuning'' aligns downstream tasks with pre-trained objectives to enable efficient knowledge transfer under limited supervision. However, current methods typically rely on single-filter backbones (e.g., low-pass), whereas real-world graphs exhibit inherent spectral diversity. Our theoretical \textit{Spectral Specificity} principle reveals that effective knowledge transfer requires alignment between pre-trained spectral filters and the intrinsic spectrum of downstream graphs.
This identifies two fundamental limitations: (1) Knowledge Bottleneck: single-filter models suffer from irreversible information loss by suppressing signals from other frequency bands (e.g., high-frequency); (2) Utilization Bottleneck: spectral mismatches between pre-trained filters and downstream spectra lead to significant underutilization of pre-trained knowledge.
To bridge this gap, we propose HS-GPPT. We utilize a hybrid spectral backbone to construct an abundant knowledge basis. Crucially, we introduce Spectral-Aligned Prompt Tuning to actively align the downstream graph's spectrum with diverse pre-trained filters, facilitating comprehensive knowledge utilization across both homophily and heterophily. Extensive experiments validate the effectiveness under both transductive and inductive learning settings.

\end{abstract}

\section{Introduction}
Graph “pre‑training and prompt tuning” \cite{sun2022gppt,liu2023graphprompt,fang2024universal,sun2023all,yu2024non} has recently emerged as a powerful paradigm for handling data scarcity in graph‑based learning. By freezing a pre‑trained Graph Neural Network (GNN) and injecting small, learnable prompt structures (e.g., virtual nodes, subgraphs, or modified edges) into the input, one can adapt the model pretrained on graphs using self-supervision to new tasks on a target graph without updating its full parameter set. Under limited supervision, these prompts effectively bridge the gap between downstream objectives and the original pre‑training task, boosting knowledge transfer. 
Despite initial success, existing prompt tuning methods rely on the homophily assumption, learning node similarities in pre-training to empower downstream tasks. However, real-world graphs often contain heterophilic edges, meaning connections between dissimilar nodes, which leads to low homophily. Recent work \cite{yu2024non} shows that homophily-based prompt tuning methods struggle to generalize across graphs with varying homophily levels, under both transductive and inductive settings.

\begin{figure}[t]
    \centering
    \subfigure[$s_{high}$ of CSBM datasets with varying homophily levels.]{
    \includegraphics[scale=0.23]{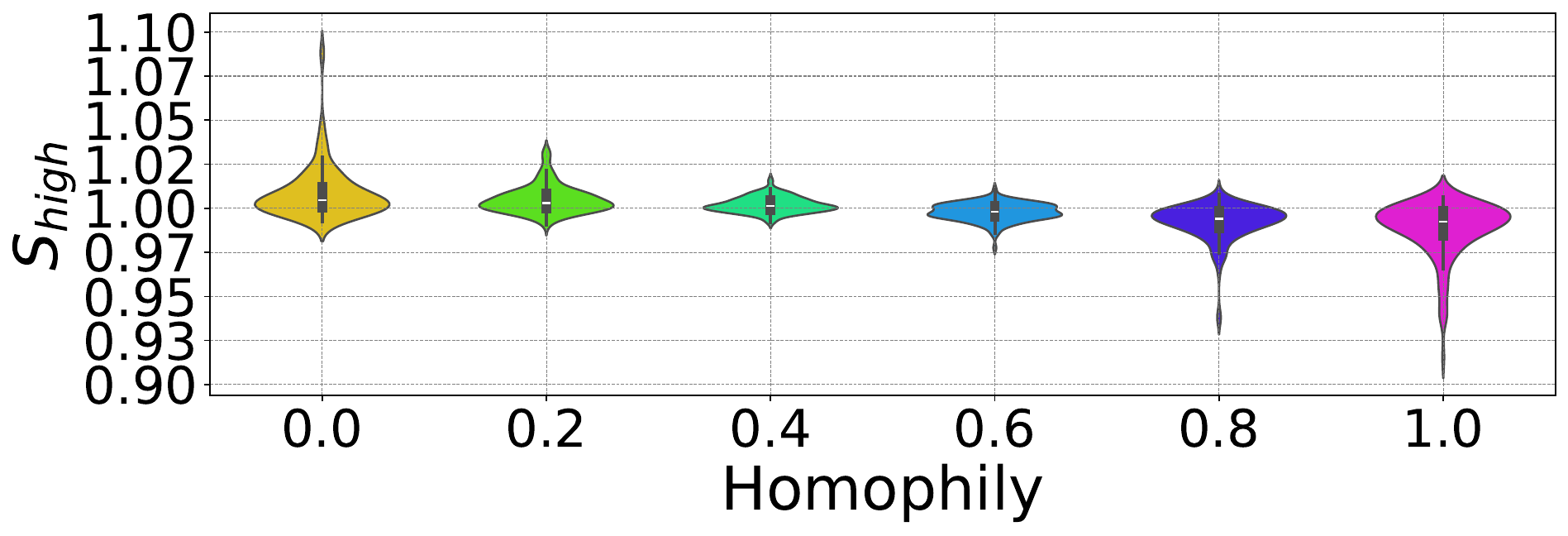}
    \label{figs:s_high_csbm}
    }
    \vskip -1pt
    \subfigure[$S_{high}$ of real-world datasets with varying homophily levels. Full names listed left to right: Cora, Citeseer, Pubmed, Texas, Cornell, Wisconsin, Chameleon, and Squirrel. The first three datasets (Cora, Citeseer, Pubmed) are considered as homophilic graphs, while the latter five are heterophilic graphs.]{
    \includegraphics[scale=0.23]{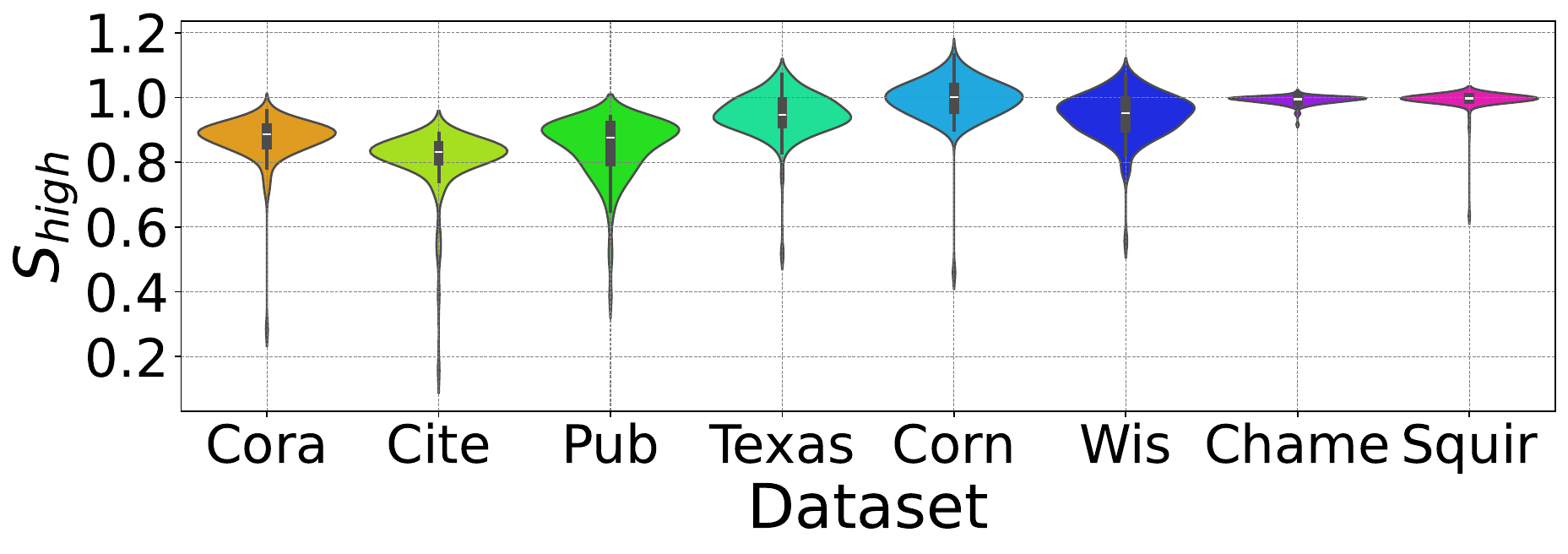}
    \label{figs:s_high_realworld}
    }
    \vskip -1em
    \caption{Distribution of $S_{high}$ (high-frequency area) across different feature dimensions in various datasets.}
    \label{fig:s_high_violin}

\end{figure}

To address this issue, we leverage the connection between homophily and spectral characteristics \cite{chen2024polygcl,pmlr-v235-wan24g,duanunifying}, and examine limitations of existing graph prompt tuning methods from a spectral perspective. As shown in Figure \ref{fig:s_high_violin}, we observe significant spectral distribution diversity across graphs with different homophily levels. Notably, lower homophily correlates with an increase in the high-frequency area $S_{high}$. 
However, current methods are typically restricted to a single kind of spectral knowledge (i.e., low-frequency information) with low-pass filters (e.g., GCN \cite{kipf2016semi}).
Given this observation, we theoretically prove the \textit{Spectral Specificity} principle: effective knowledge transfer requires alignment between pre-trained spectral filters and the intrinsic spectrum of downstream graphs.
This reveals a twofold failure in existing methods: (1) \textbf{Knowledge Bottleneck} in the pre-training stage: single-filter models (e.g., low-pass filters) physically suppress high-frequency signals fundamental to heterophilic tasks, causing irreversible information loss; (2) \textbf{Utilization Bottleneck} in the prompt-tuning stage: even with a hybrid backbone, the inherent spectral diversity of downstream graphs often leads to partial alignment. Since each pre-trained filter captures a distinct frequency band, a specific downstream graph typically aligns with only a limited subset of filters, leaving the rich knowledge encoded in the remaining non-aligned filters underutilized.

To address the heterophily issue in prompt tuning, we propose the \textbf{H}ybrid \textbf{S}pectral \textbf{G}raph \textbf{P}re-training and \textbf{P}rompt \textbf{T}uning model (HS-GPPT), which overcomes two challenges. 
The first challenge is: \textit{What spectral knowledge should be learned, and how?} 
Our analysis shows that acquiring abundant spectral knowledge is essential for generalization. Therefore, we adopt a hybrid GNN backbone, where each filter captures a distinct spectral band. Building on prior methods \cite{chen2024polygcl} that handle only low- and high-pass filters, we extend the pre-training process to cover diverse frequency bands, constructing an abundant spectral basis.

The second challenge is: \textit{How can prompts adaptively align with this abundant spectral knowledge to ensure full utilization?}
We introduce spectral-aligned prompt graphs: lightweight, learnable subgraphs that actively reshape the spectral distribution of downstream graphs to match the spectrum of the pre-trained filters. 
We theoretically prove the \textit{Spectral Adaptability} of our prompts: for each frozen filter, a dedicated prompt can be optimized to transform the input spectrum. This strategy ensures that the abundant pre-trained spectral knowledge is fully utilized, maximizing the power of the frozen backbone.

In summary, our contributions are: (i) The first theoretical analysis of hybrid spectral graph prompt tuning, proving the need for rich spectral pre-training and establishing spectral alignment principles; (ii) HS-GPPT, a novel framework leveraging abundant spectral knowledge for spectral downstream alignment, boosting knowledge transfer; (iii) Experiments across datasets with varying homophily show HS-GPPT outperforms baselines in both transductive and inductive settings, validating its generalization.

\section{Related Work}
We briefly introduce the related work here, while more details are in Appendix \ref{appendix:related_work}.

\noindent\textbf{Graph Prompt Tuning}. Graph prompt tuning \cite{sun2022gppt,fang2024universal,sun2023all,liu2023graphprompt,yu2024non} tackles data scarcity by leveraging frozen pre-trained knowledge through prompts. For example, GPrompt \cite{liu2023graphprompt} uses prompt vectors to unify pre-training and downstream tasks under a common template, while GPF \cite{fang2024universal} inserts prompt nodes adaptable to various pre-training strategies. These methods, tailored to homophilic graphs, rely mainly on low-frequency signals and can be seen as special cases of our approach when restricted to low-frequency information. Our spectral analysis shows that under sparse supervision, large spectral gaps impede effective parameter learning, leading to the failure of existing homophily-based methods. 
Although ProNoG \cite{yu2024non}, HeterGP \cite{yan2025hetergp}, and DAGPrompT \cite{chen2025dagprompt} have recently extended prompt tuning to heterophilic graphs via distribution-aware strategies, they operate in the spatial domain and rely on heuristic adaptations.
Crucially, they overlook the fundamental spectral mismatch between the frozen backbone and diverse downstream graphs. In contrast, our method treats prompt tuning as a spectral alignment process, actively aligning the input spectrum to match pre-trained filters for better performance.

\noindent\textbf{Heterophilic Graph Learning}. Recent methods \cite{xiao2024simple,bo2021beyond,chen2024polygcl,pmlr-v235-wan24g}  design GNN architectures and pre-training objectives for heterophilic graphs, such as leveraging high-frequency information \cite{bo2021beyond,chen2024polygcl,pmlr-v235-wan24g} and discovering potential neighbors \cite{jin2021node,pei2020geom}. However, they rely on full-model fine-tuning. Without a mechanism to align pre-trained knowledge with downstream tasks, directly using their heterophilic priors can cause knowledge underutilization and even negative transfer, especially under limited supervision. Thus, these models and pre-training schemes for heterophilic graphs are ill-suited for prompt tuning methods.

\section{Preliminary and Theoretical Analysis}

\subsection{Preliminaries}
\noindent\textbf{Problem Formulation}. 
Let $\mathcal{G} = {\{\mathcal{V}, \mathbf{X}, {\mathcal{E}}}\}$ be a graph, where $\mathcal{V}$ denotes the set of nodes $ {\{v_1,...,v_N\}}$, $\mathcal{E}=\{e_{ij}\}$ is the set of undirected edges, and the feature matrix $\mathbf{X}\in \mathbb{R}^{N\times d}$ consists of $d$-dimensional features of $N$ nodes. In this paper, we focus on graph ``pre-training and prompt tuning'', which first trains a graph model on $\mathcal{G}$ using self-supervised tasks, then freezes the pre-trained model and adopts learnable prompts to reformulate the downstream tasks to align with the pretexts. Formally, let $F_{\mathbf{\theta}^*}$ denote a pre-trained graph model with frozen parameters $\mathbf{\theta}^*$, and $P_{\mathbf{\omega}}$ denotes a graph prompt function with learnable parameters $\mathbf{\omega}$. $\mathcal{L}_{down}$ represents the learning objective of the downstream task. The graph prompt tuning is formulated as:
\begin{equation}
    \mathbf{\omega}^* = \arg\max_{\omega}~\mathcal{L}_{down}\big(F_{\mathbf{\theta}^*}\big(P_{\mathbf{\omega}}(\mathcal{G})\big)\big).
\end{equation} 
Recognizing that node classification is the main challenge on heterophilic graphs, we follow prior work \cite{bo2021beyond,chen2024polygcl} and select this task as our downstream objective. Specifically, the graph prompt tuning occurs under limited supervision, where for each class, only $K$ nodes are provided, also known as $K$-shot learning.

\noindent\textbf{Homophily}. Homophily refers to the tendency of nodes to connect with other nodes that are similar to themselves. Edge homophily level $h$ \cite{zhu2020beyond} measures the fraction of homophilic edges that connect nodes that share the same label, i.e., $h = \frac{|\{e_{uv} \in{\mathcal{E}}: y_u = y_v\}|}{|\mathcal{E}|} \in [0,1]$, 
where $|\mathcal{E}|$ denotes the number of edges and $y_i$ is the label of node $v_i$. A larger $h$ means a larger degree of homophily.

\noindent\textbf{Spectral Graph Filters}.
Given a graph $\mathcal{G} = {\{\mathcal{V}, \mathbf{X}, {\mathcal{E}}}\}$, let $\mathbf{A} \in \mathbb{R}^{N\times N}$ be the adjacency matrix and $\mathbf{D} \in \mathbb{R}^{N\times N}$ denote the diagonal degree matrix with $\mathbf{D}_{ii}=\sum_{j}\mathbf{A}_{ij}$. 
The graph Laplacian matrix $\mathbf{L}$ is $\mathbf{D}-\mathbf{A}$ (regular) or $\mathbf{I}-\mathbf{D}^{-\frac{1}{2}}\mathbf{A}\mathbf{D}^{-\frac{1}{2}}$ (normalized), where $\mathbf{I}$ is an identity matrix. 
$\mathbf{L}$ can be eigendecomposed as $\mathbf{L}=\mathbf{U}\mathbf{\Lambda}\mathbf{U}^T$, where $\mathbf{U}=[\mathbf{u}_1,\mathbf{u}_2,...,\mathbf{u}_N]$ are orthonormal eigenvectors and $\mathbf{\Lambda}=diag([\lambda_1,\lambda_2,...,\lambda_N)]$ are corresponding eigenvalues. Typically, $\mathbf{U}$ is the graph Fourier basis and $\mathbf{\Lambda}$ are the frequencies. The objective of spectral graph filters is to design a function $g(\cdot)$ on $\mathbf{\Lambda}$ to learn the graph representation $\mathbf{Z}$:
\begin{equation}
    \mathbf{Z}=g(\mathbf{L})\mathbf{X}=\mathbf{U}g(\mathbf{\Lambda})\mathbf{U}^T\mathbf{X}.
\end{equation}
From a spatial view, different filtering characteristics aggregate diverse neighbor information. Low-pass filters capture node-neighbor similarity, while high-pass filters capture differences \cite{luo2024spectral}.

Suppose the feature dimension $d=1$, the nodes' feature is denoted as $\mathbf{x}\in\mathbb{R}^{N\times 1}$. The spectral graph signal $\mathbf{\hat{x}}$ is obtained by $\mathbf{\hat{x}}=(\hat{x}_{1},\hat{x}_{2},...,\hat{x}_{N})^T = \mathbf{U}^T\mathbf{x}$, where $\hat{x}_{i} = \mathbf{u}_i^T \mathbf{x}$ gives the projection of $\mathbf{x}$ to frequency $\lambda_i$. We denote the spectral energy as $\hat{x}_k^2/\sum_{i=1}^N\hat{x}_i^2$, which reflects the spectral distribution. The larger the $\hat{x}_k^2/\sum_{i=1}^N\hat{x}_i^2$ is, the more frequency components corresponding to \(\lambda_i\) will be. To observe the spectral distribution intuitively, we introduce the high-frequency area ~\cite{tang2022rethinking}.

\begin{definition}[High-frequency Area]
For a graph with spectral signal $\mathbf{\hat{x}}=(\hat{x}_{1},\hat{x}_{2},...,\hat{x}_{N})^T = \mathbf{U}^T\mathbf{x}$, the high-frequency area is calculated as $S_{high}=\frac{\sum_{k=1}^N\lambda_k\hat{x}_k^2}{\sum_{k=1}^N\hat{x}_k^2}=\frac{\mathbf{x}^T\mathbf{L}\mathbf{x}}{\mathbf{x}^T\mathbf{x}}$. $S_{high}$ quantifies the spectral distribution ($S_{high} \in [0, 2]$). A larger $S_{high}$ indicates stronger high-frequency components, while a smaller implies greater low-frequency components.
\end{definition}

This metric provides a unified characterization of both structure $\mathbf{L}$ and nodal attributes $\mathbf{x}$. As $S_{high}$ is obtained from each single dimension, for simplicity, the theoretical analysis is also in a unidimensional setting, while the conclusions also hold in the multidimensional case.

\subsection{Theoretical Analysis}
\label{section:thero_movatation}

In this section, we build the theoretical motivation for HS-GPPT. We first quantify the \textit{spectral distribution diversity}, proving that structural shifts (e.g., homophily changes) inherently lead to significant spectral shifts. 
We then establish the \textit{spectral specificity} principle: effective knowledge transfer requires strict alignment between pre-trained spectral filters and the downstream graph's intrinsic spectrum.
This highlights a dual necessity: constructing an abundant spectral basis (via pre-training) and employing spectral alignment (via prompt tuning) to ensure this knowledge is fully utilized.

\subsubsection{Spectral Distribution Diversity}
\label{sec:spectral_diversity}

To understand the challenge of transferring knowledge across graphs, we first verify the connection between graph structure (homophily) and spectral properties. 
Given a graph $\mathcal{G}$ with edge set $\mathcal{E}$ and homophily level $h$, we define the expected normalized intra-class and inter-class distances as $\mathbb{E}(d_{intra})$ and $\mathbb{E}(d_{inter})$, respectively \cite{ju2025cluster} (details in Appendix \ref{section:appendix_spectrum_dis}). 
Generally, connected nodes within the same class tend to have similar features, implying $\mathbb{E}(d_{inter}) > \mathbb{E}(d_{intra})$.
Based on the definition of high-frequency area $S_{high}$, we formalize the spectral-homophily correlation in Proposition \ref{prop:spectrum_dis}.

\begin{proposition}[Homophily-Spectrum Correlation]
\label{prop:spectrum_dis}
The spectral energy distribution ($S_{high}$) is related to the homophily level $h$. Assuming bounded feature distances, $S_{high}$ exhibits a linear negative correlation with $h$:
\begin{equation}
    S_{high} = \mathcal{C}_{base} - \mathcal{C}_{gap} \cdot h,
    \label{eq:linear_relation}
\end{equation}
where $\mathcal{C}_{base} = {|\mathcal{E}|}\mathbb{E}(d_{inter})$ and $\mathcal{C}_{gap} = {|\mathcal{E}|}[\mathbb{E}(d_{inter})-\mathbb{E}(d_{intra})]$ are positive constants derived from the graph's edge count and feature statistics.
\end{proposition}

The proof is provided in Appendix \ref{section:appendix_spectrum_dis}. Proposition \ref{prop:spectrum_dis} reveals that spectral distributions are closely tied to homophily: a decrease in $h$ inherently leads to an increase in spectral energy $S_{high}$. This confirms the existence of spectral distribution diversity among downstream graphs with different homophily levels. We also validate this diversity using both synthetic Contextual Stochastic Block Models (CSBM) and real-world datasets. 
As shown in Figure \ref{figs:s_high_csbm} (details in Appendix \ref{section:appendix_csbm}), as $h$ varies from 0 to 1, the spectral distributions (Violin Plots) exhibit clear diversity, perfectly aligning with the trend predicted in Eq. \ref{eq:linear_relation}.
Similarly, real-world datasets (Figure \ref{figs:s_high_realworld}) show distinct spectral patterns: homophilic graphs (e.g., Cora \cite{mccallum2000automating}, Citeseer \cite{sen2008collective}) are dominated by low-frequency components, while heterophilic ones (e.g., Texas \cite{pei2020geom}, Chameleon \cite{rozemberczki2021multi}) show strong high-frequency energy.

This diversity poses a fundamental challenge: since standard pre-trained models are typically restricted to a single frequency band (mostly low-pass), transferring to diverse downstream tasks creates a significant \textbf{spectral gap}.

\begin{figure*}
    \centering
    \includegraphics[scale=0.13]{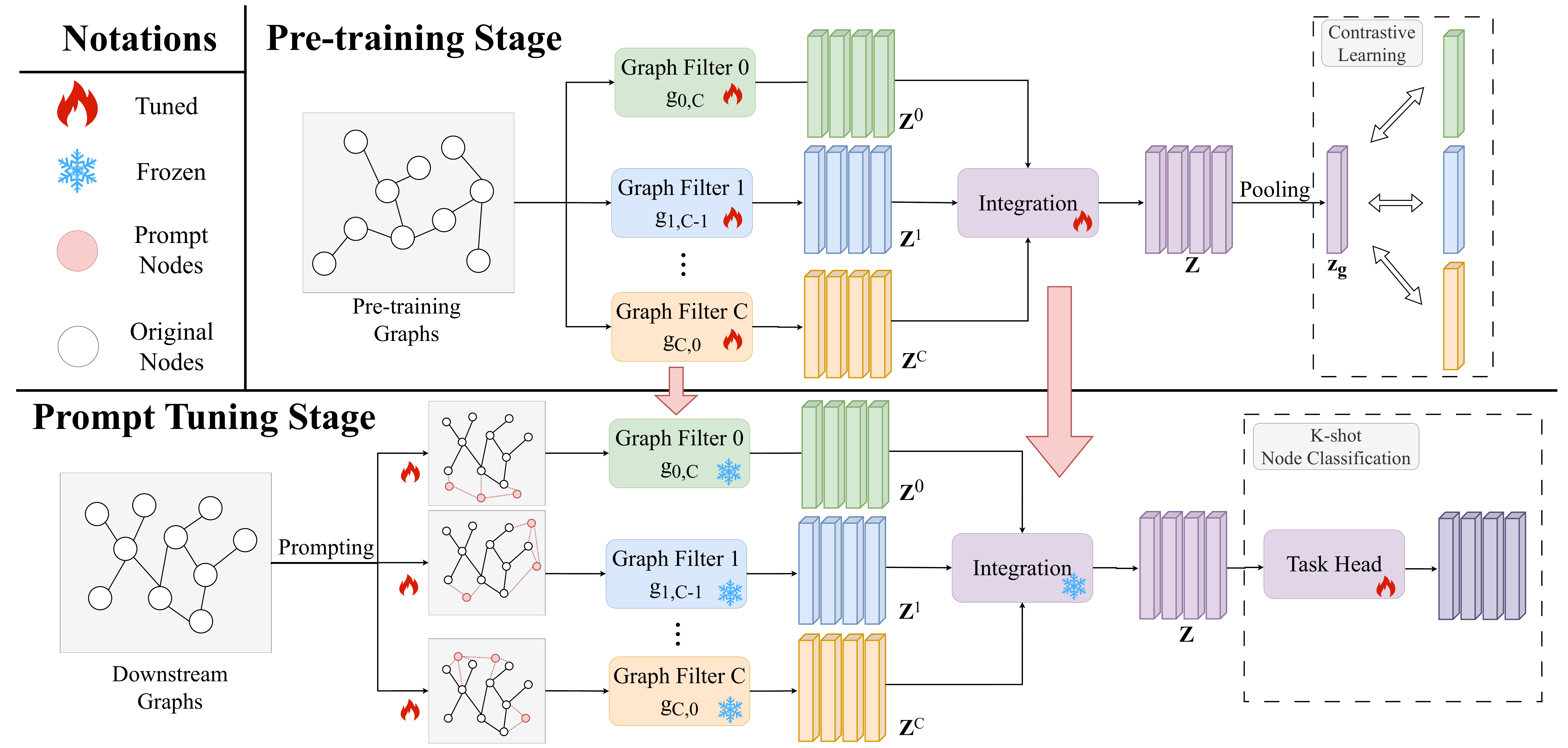}
    \vskip -1em
    \caption{The overall framework of our HS-GPPT. In the pre-training stage, the graph filters and integration weights are trained. In the prompt tuning stage, we keep the graph filters and integration weights frozen and only tune the learnable prompt graphs and task head (i.e., one-layer MLP).}
    \label{figs:framework}
    \vskip -1em
\end{figure*}

\subsubsection{Spectral Specificity and Generalization Limitation}
\label{section:spe_spe}

Having established the spectral distribution diversity, we now investigate how this diversity impacts GNN generalization. We aim to prove the \textit{Spectral Specificity} principle: a graph filter performs optimally only when its spectral response matches the input graph's spectrum.

We employ the Spectral Regression Loss (SRL) \cite{lei2022evennet} to evaluate the filter $g$. Minimizing SRL serves as a proxy for maximizing downstream performance.

\begin{theorem}[Spectral Specificity]
\label{theorem:spe_spe}
Given a pre-trained graph filter $g$ and an input graph $\mathcal{G}$ with spectral signal energy $\mathbf{\hat{x}}=(\hat{x}_{1},\ldots,\hat{x}_{N})^T$, to minimize the upper bound of the SRL, the filter response $g(\lambda_i)$ must be positively correlated with the signal energy $\hat{x}_i^2$. 
Specifically, the filter should assign larger amplification $g(\lambda_i)$ to frequencies where the signal energy $\hat{x}_i^2$ is larger.
\end{theorem}

The detail of SRL and proof are provided in Appendix \ref{section:appendix_of_spe_spe}. Theorem \ref{theorem:spe_spe} confirms that optimal performance requires alignment between pre-trained spectral filters and the downstream graph’s intrinsic spectrum. We verify the claim by testing spectral graph filters (i.e., low/mid/high-pass) on CSBM datasets from Figure \ref{figs:s_high_csbm}. Results (details in Appendix \ref{section:appendix_triple_gnn}) show distinct graph filters excel on different graph types, aligning with our theorem. This imposes two fundamental requirements.

First, \textbf{the pre-trained spectral knowledge must be abundant}. Since downstream spectral distributions vary widely, a single-filter backbone suffers from irreversible information loss when signals fall outside its fixed passband. Thus, a hybrid spectral backbone is essential to ensure an abundant knowledge basis.

Second, and more crucially, \textbf{this abundant knowledge must be fully utilized}. Since each pre-trained filter captures a distinct frequency band, a specific downstream graph typically aligns with only a limited subset of filters, leaving the rich knowledge encoded in the remaining non-aligned filters underutilized. This dilemma motivates our method, which employs prompt tuning to actively transform the input spectrum, ensuring the abundant frozen spectral knowledge is fully utilized.

\section{Methodology}
\label{sec:method}

Based on the theoretical analysis in Section \ref{section:thero_movatation}, we propose HS-GPPT. The framework consists of two stages: (1) \textbf{Hybrid Spectral Pre-training}, where we construct a backbone (Section \ref{section:hybrid_backbone}) to capture comprehensive spectral knowledge (Section \ref{section:pre-training}); and (2) \textbf{Spectral-Aligned Prompt Tuning}, where we align downstream graphs with the pre-trained backbone to enhance the knowledge transfer (Section \ref{section:prompt_tuning}). The overall framework is in Figure \ref{figs:framework}, Algorithm \ref{alg:pretraining_algorithm}, and Algorithm \ref{alg:prompt_algorithm} in the appendix.

\subsection{Hybrid Spectral Filter Backbone}
\label{section:hybrid_backbone}

To generalize across graphs with varying spectral distributions, we require a backbone capable of capturing information from a wide range of frequencies. We adopt the Beta Wavelet GNN (BWGNN) \cite{tang2022rethinking} as our backbone implementation due to its flexible band-pass properties, where each focuses on different frequency components.
The graph filters are formulated as:
\begin{equation} \small
    g_{k,r}(\mathbf{L})=\frac{{\frac{\mathbf{L}}{2}}^k(\mathbf{I}-\frac{\mathbf{L}}{2})^{r}}{2B(k+1,r+1)},
\end{equation}
where $B(k+1,r+1)=\frac{k!r!}{(k+r+1)!}$ is the Beta constant. With different constant values of $k$ and $r$, $g_{k,r}$ possesses different spectral characteristics. By setting $k + r = C$ as a constant, we obtain a group of filters  $g_{\beta}=(g_{0,C},g_{1,C-1},\cdots,g_{C,0})$. Here, $g_{k,C-k}$ acts as a specific spectral filter. More details and visualization of BWGNN are in Appendix \ref{section:appendix_bwgnn} and Figure \ref{figs:bwgnn}. We also evaluate the performance of other hybrid GNN backbones in Section \ref{section:compat_invest}.
With these filters, we extract spectral representations from multiple filters:
\begin{equation}
    \label{equation:tem_1}
    \mathbf{Z}^k = \text{MLP}_k\Big( g_{k,C-k}(\mathbf{L})\mathbf{X} \Big), \quad k \in \{0, \dots, C\}.
\end{equation}
Here the $i$-th row of $\mathbf{Z}^k$, i.e., $\mathbf{z}_i^k$, is the representation of node $v_i$ under the $k$-th filter.

\subsection{Hybrid Spectral Pre-training}
\label{section:pre-training}

The hybrid backbone generates diverse spectral representations $\mathbf{Z}^k$. To initialize these filters and learn their effective integration, we employ self-supervised contrastive learning. Building on previous work \cite{chen2024polygcl}, which uses local-global contrastive learning \cite{velickovic2019deep} to align low- and high-frequency views, we extend this to multiple spectral views. We perform contrastive learning between node embeddings in different views (as local patches) and integrated graph embeddings (as global summaries), as shown in the upper part of Figure \ref{figs:framework}.

To derive the global summary, we first integrate node representations from different spectral views. Since the spectral distribution diversity across feature dimensions, we employ an element-wise weighted sum to fuse the views adaptively:
\begin{equation} \small
    \label{equation:pre_inte}
     \mathbf{z}_i = \sum_{k=0}^C \frac{\exp(\mathbf{w}^k)}{\sum_{j=0}^C \exp(\mathbf{w}^j)}\odot \mathbf{z}_i^k,
\end{equation}
where $\odot$ denotes the element-wise product and $\mathbf{w}^k$ is the learnable integration weight vector. We then apply mean pooling over all nodes to obtain the global graph summary $\mathbf{z_{g}} =\frac{1}{N}\sum_{i=1}^N\mathbf{z}_i$.

For the contrastive objective, we generate negative samples by corrupting the graph structure (i.e., shuffling nodes while preserving the adjacency matrix), yielding negative embeddings $\mathbf{z}_{i}^{k-}$.
We use a discriminator $\mathcal{D}$ to measure the agreement between node and graph embeddings $\mathcal{D}(\mathbf{z}_i^k,\mathbf{z}_{g})=\sigma(\mathbf{z}_i^k\mathbf{W}\mathbf{z}_{g}^T) \in [0,1]$, where $\mathbf{W}\in \mathbb{R}^{d\times d}$ denotes the weight matrix and $\sigma$ denotes the sigmoid activation function. The multi-view local-global mutual information maximization objective is formulated as:
\begin{equation} \small
    \label{equation:pre_loss}
    \mathcal{L}_{pre}(\mathbf{\theta})=-\frac{1}{CN}\sum_{k=0}^C\sum_{i=1}^N\Big(\log \mathcal{D}(\mathbf{z}_i^k,\mathbf{z}_{g})+\log(1 - \mathcal{D}(\mathbf{z}_i^{k-},\mathbf{z}_{g}))\Big).
\end{equation}
The learnable parameters $\mathbf{\theta}$ include MLP parameters in each graph filter, integration weight vectors $\mathbf{w}^k$ for $k=(0,1,...,C)$, and the weight matrix of discriminator $\mathcal{D}$.
Through this process, the model captures comprehensive spectral knowledge and learns to integrate it effectively. After pre-training, these parameters are fixed to serve as a stable spectral basis for the subsequent prompt tuning.

\subsection{Spectral-Aligned Prompt Tuning}
\label{section:prompt_tuning}

During the pre-training stage, the backbone learns a diverse set of graph filters and their integration schemes, learning rich spectral knowledge across different frequency bands. 
However, a specific downstream graph often exhibits a specific spectral bandwidth. According to the \textit{Spectral Specificity} principle (Theorem \ref{theorem:spe_spe}), this mismatch leads to knowledge underutilization: only the filters matching the graph's dominant frequency are fully utilized, while other valuable filters remain underutilized. 

To address this, we propose \textbf{Spectral-Aligned Prompt Tuning}. Instead of modifying the frozen model, we optimize the input graph structure to align with each pre-trained filter individually. By transforming the graph's spectral distribution to match the spectrum of every filter, we ensure that \textbf{all pre-trained spectral knowledge is effectively activated}, maximizing knowledge transfer (see Figure \ref{figs:framework}, lower part).

\noindent\textbf{Prompt Construction.}
We design the prompt as a graph $\mathcal{G}_p=(\mathcal{V}_p, \mathbf{P}, \mathcal{E}_p)$. Here $\mathcal{V}_p$ is the prompt node set and $\mathbf{P} \in \mathbb{R}^{N_p\times d}$ are learnable node representations, where $N_p$ is the number of prompt nodes and $d$ is the feature dimension matching that of original graph nodes. Edges $\mathcal{E}_p$ are constructed based on the similarity between prompt nodes $e_{ij} = \mathbb{I}\big({\sigma(\mathbf{p}_i\cdot \mathbf{p}_j^T) > \tau_{inner}}\big)$, where $\mathbb{I}$ is the indicator function and $\tau_{inner}$ is the pre-defined threshold for inner edges within the prompt graph. $\mathbb{I}\big({\sigma(\mathbf{p}_i\cdot \mathbf{p}_j^T) > \tau_{inner}}\big)$ outputs 1 if $\sigma(\mathbf{p}_i\cdot \mathbf{p}_j^T)$ is larger than $\tau_{inner}$; otherwise it outputs 0.

The prompt $\mathcal{G}_P$ will then be inserted to the original graph $\mathcal{G}$ to align their spectral patterns with those of pre-trained spectral filters. As the feature distribution and scale of $\mathcal{G}_P$ and $\mathcal{G}$ are different, to avoid introducing noise, we first normalize the prompt graphs' node features to match the distribution of the original graph $\mathcal{G}$ as:
\begin{equation} \small
    \label{equation:prompt_norm}
    \mathbf{p}_i^{\prime} = \frac{\mathbf{p}_i-\mu_{p}}{\sigma_{p}}\sigma_{o}+\mu_{o},
\end{equation}
where $\mu_p,\sigma_p$, $\mu_o, \sigma_o$ are the mean and standard deviation of the prompt and original graphs, respectively. We insert edges between nodes in $\mathcal{G}_p$ and $\mathcal{G}$ to attach the prompt graph into the whole original graph by $e_{ij} = \mathbb{I}\big({\sigma(\mathbf{p}_i^{\prime}\cdot \mathbf{x}_j^T) > \tau_{cross}}\big)$, where $\tau_{cross}$ denotes the pre-defined threshold for cross edges. Adjusting $\tau_{cross}$ controls node similarity, enabling selective introduction of diverse frequency components. The final prompted graph  $\tilde{\mathcal{G}}=\psi(\mathcal{G},\mathcal{G}_p)$, where $\psi$ denotes the insertion strategy.

\begin{table*}
\centering
\scalebox{0.84}{
 \begin{tabular}{c|c|c|c|c|c|c|c|c|c|c}
 \toprule
 Method   & Cora & Pubmed & Citeseer & Cornell  & Texas & Wisconsin & Chameleon & Squirrel & Ratings & Empire \\ \cline{1-11} \hline \hline              
                        GCN     & 0.6506 & 0.5405 & 0.4187 & 0.1835 & 0.2506 & 0.2496 & 0.2998 & 0.2441 & 0.1911 & 0.1491 \\
                        GAT & 0.6091 & 0.5381 & 0.4162 & 0.1531 & 0.2144 & 0.1836 & 0.2873 & 0.2281 & 0.1827 & 0.1367  \\
                        TFE-GNN & 0.3286 & 0.4590 & 0.2874  & \underline{0.3950} & 0.3730 & \underline{0.3828} & 0.2872 & 0.2242 & 0.1939 & \underline{0.3155} \\
                        \cline{1-11}
                        DGI & 0.3187 & 0.3743 & 0.2235 & 0.1650 & 0.3031  & 0.2293  & 0.2692 & 0.1989 & 0.1485 & 0.0710 \\
                        GraphCL  & 0.5603 & 0.5576 & 0.3741 & 0.1826 & 0.2673 & 0.2414  & 0.2710 & 0.1890 & 0.1853 & 0.0961 \\ 
                        SimGRACE & 0.4283 & 0.4316 & 0.3412 & 0.1904 & 0.2833 & 0.2116 & 0.2706 & 0.2071 & 0.1761 & 0.0635\\
                        PolyGCL & \underline{0.6655} & \underline{0.6782} & \underline{0.5010} & 0.2268 & \underline{0.4913} & 0.2254 & \underline{0.3308} & \underline{0.2450} & 0.1874 & 0.0528 \\
                        \cline{1-11}
                        GPPT & 0.5109 & 0.6187 & 0.3731 & 0.1529 & 0.2702 & 0.1851 & 0.2933 & 0.2105 & \underline{0.1947} & 0.0666 \\
                        Gprompt  & 0.5011 & 0.5452 & 0.4149 & 0.1710 & 0.1755 & 0.1938 & 0.2258 & 0.2015 & 0.1735 & 0.0522 \\
                        GPF-plus  & 0.5721 & 0.5932 & 0.3534 & 0.1891 & 0.2376 & 0.0926 & 0.1976 & 0.1703 & 0.1833 & 0.0383 \\
                        All-in-One  & 0.3648 & 0.4141 & 0.1775 & 0.1341 & 0.1492 & 0.1174 & 0.2254 & 0.1751 & 0.1819 & 0.0275 \\
                        ProNoG  & 0.5564 & 0.5242 & 0.2466 & 0.1987 & 0.2627 & 0.2218 & 0.2565 & 0.1890  & 0.1963 & 0.0784 \\
                        \cline{1-11}
                        HS-GPPT  & \textbf{0.6915} & \textbf{0.6910} & \textbf{0.5043} & \textbf{0.4209} & \textbf{0.5724} & \textbf{0.4554} & \textbf{0.3324} & \textbf{0.2536} & \textbf{0.1972} & \textbf{0.3520} \\ \bottomrule
 \end{tabular}}
 \caption{Performance comparison on 5-shot node classification under the transductive setting. Bold signifies the best result across all methods, while underline highlights the best baseline result. Additional results (e.g., standard deviations, baselines with other backbones) are in Appendix \ref{section:appendix_acc} and \ref{section:appendix_graphtransformer}.}
\label{tab:perform_transductive}
\vskip -1em
\end{table*}

\begin{table}
\centering
\scalebox{0.8}{
 \begin{tabular}{c|c|c|c|c}
 \toprule
  Settings & \multicolumn{2}{c|}{In-Domain} & \multicolumn{2}{c}{Cross-Domain} \\ \cline{1-5}
  Source & Wisconsin& Chameleon & Pubmed   & Squirrel  \\  
 Target & Texas  & Squirrel & Texas  & Cornell  \\  \hline \hline        
                        DGI & 0.1976  & 0.2048 & 0.2177   & 0.1450   \\
                        GraphCL  & 0.2847 & 0.2098 & 0.2684  & 0.1686   \\ 
                        SimGRACE & 0.2309  & 0.2111 & 0.1834  & 0.1369 \\
                        PolyGCL & \underline{0.3901}  & \underline{0.2130} & 0.2934  & \underline{0.2112}  \\
                        \cline{1-5}
                        GPPT & 0.2675  & 0.1957 & \underline{0.3026}  & 0.1443 \\
                        Gprompt  & 0.1677  & 0.1926 & 0.1637  & 0.1727 \\
                        GPF-plus  & 0.1858  & 0.1702 & 0.2470  & 0.1613  \\
                        All-in-One  & 0.1321  & 0.1752 & 0.1393  & 0.1224  \\
                        ProNoG  & 0.1878  & 0.1868 & 0.2155 & 0.1869  \\
                        \cline{1-5}
                        HS-GPPT  & \textbf{0.4428} & \textbf{0.2307} & \textbf{0.5106}  & \textbf{0.4247}  \\ \bottomrule
 \end{tabular}}
 \caption{Performance comparison on 5-Shot node classification under the inductive setting. Here `Source' denotes the pre-training datasets and `Target' denotes the downstream dataset. More results are in Appendix \ref{section:appendix_acc} and \ref{section:appendix_graphtransformer}.}
\label{tab:perform_inductive}
\vskip -1em
\end{table}
\noindent\textbf{Theoretical Guarantee.}
To theoretically justify our approach, we provide Theorem \ref{theorem:spec_ada}, which guarantees the universal adaptability of the prompted graph.

\begin{theorem}[Spectral Adaptability]
    \label{theorem:spec_ada}
    Given a pre-trained model $F$, along with two graphs $\mathcal{G}_1$ and $\mathcal{G}_2$, assume their spectral distributions differ, i.e., $S_{high1} \neq S_{high2}$. Then, there exists a prompt graph $\mathcal{G}_p$ such that $F\big(\psi(\mathcal{G}_1,\mathcal{G}_p)\big)=F(\mathcal{G}_2)$, where $\psi$ denotes the insertion strategy.
\end{theorem}
Proof in Appendix \ref{appendix:spec_ada}. Theorem \ref{theorem:spec_ada} indicates that the prompted graph can theoretically approximate graphs of any spectral distribution. Combining this capability with the specificity requirement (Theorem \ref{theorem:spe_spe}), we derive the following alignment guarantee:

\begin{corollary}[Spectral Alignment]
    \label{corollary:srl_bound}
     Given a pre-trained model $F$, we denote $\mathcal{L}_{UB}$ as the upper bound of SRL on downstream tasks. For an input downstream graph $\mathcal{G}$, there exists a prompt graph $\mathcal{G}_p$ that satisfies: 
    $\mathcal{L}_{UB}\big(\psi(\mathcal{G},\mathcal{G}_p)\big)\leq\mathcal{L}_{UB}(\mathcal{G})$.
\end{corollary}

The proof is in Appendix \ref{appendix:corollary}. Corollary \ref{corollary:srl_bound} confirms that using a proper prompt enhances performance by aligning the downstream graph's spectrum with pre-trained knowledge.

\noindent\textbf{Filter-Specific Alignment Strategy.}
To fully utilize the backbone, we cannot rely on a single global prompt, as it might fail to align with the disjoint passbands of the $C+1$ filters simultaneously.
Therefore, we employ a fine-grained filter-specific alignment strategy. For each graph filter $g_{k,C-k}$ in the backbone, we assign a dedicated prompt graph $\mathcal{G}_p^k=(\mathcal{V}_p^k,\mathbf{P}^k,\mathcal{E}_p^k)$. 
This generates independent prompted graphs $\tilde{\mathcal{G}}^k=\psi(\mathcal{G},\mathcal{G}_p^k)$, each optimized to activate a specific spectral view. The representations are then computed and integrated:
\begin{equation} \small
    \label{equation:prompt_agg_inte}
    \mathbf{\tilde{Z}}^k=\text{MLP}_k\Big(g_{k,C-k}(\tilde{\mathbf{L}}^k)\mathbf{\tilde{X}}^k\Big), \quad
     \mathbf{\tilde{z}}_i = \sum_{k=0}^C \frac{\exp(\mathbf{w}^k)}{\sum_{j=0}^C \exp(\mathbf{w}^j)}\odot \mathbf{\tilde{z}}_i^k,
\end{equation}
where $\tilde{\mathbf{L}}^k$ and $\mathbf{\tilde{x}}_i^k$ are the Laplacian matrix and node features in the prompted graph $\tilde{\mathcal{G}}^k$. Here both MLP and $\mathbf{w}^k$ are the frozen pre-trained parameters. By aligning the downstream graph with the specific spectral characteristics of each filter, the model ensures that all pre-trained knowledge is fully used.

\noindent\textbf{Model Optimization.}
The integrated representation $\tilde{\mathbf{z}}_i$ is fed into a lightweight task head (one-layer MLP) for classification. The model is optimized via the cross-entropy loss:
\begin{gather} \small
    \label{equation:prompt_loss} 
    \mathbf{l}_i=\text{Softmax}\big(\text{MLP}_{head}(\tilde{\mathbf{z}}_i)\big), \\
    \mathcal{L}_{down}(\omega;\theta^*) = -\frac{1}{N}\sum_{i = 1}^{N}\log(\mathbf{l}_{i})\mathbf{y}_{i}^T,
\end{gather}
where $\mathbf{y}_i$ is the one-hot label vector. Note that $\theta^*$ represents the frozen pre-trained parameters (including filter MLPs and integration weights $\mathbf{w}^k$), while $\omega$ denotes the learnable parameters (prompt features $\mathbf{P}^k$ and task head).

We discuss the parameter and time complexity in Appendix \ref{section_appendix_complexity}. Compared with a $(C+1)$-layer GCN, where $C+1$ is the number of graph filters in our backbone, our method introduces minimal additional parameters, ensuring efficiency.

\section{Experiments}

\subsection{Experimental Setup}

\noindent\textbf{Datasets}. We conduct experiments on real-world datasets with various homophily levels. Among them, \textit{Cora} \cite{mccallum2000automating}, \textit{Citeseer} \cite{sen2008collective}, and \textit{Pubmed} \cite{kipf2016semi} are homophilic graphs, \textit{Cornell}, \textit{Texas}, \textit{Wisconsin} \cite{pei2020geom}, \textit{Chameleon}, \textit{Squirrel} \cite{rozemberczki2021multi}, \textit{Amazon-ratings} (\textit{Ratings}) and \textit{Roman-empire} (\textit{Empire}) \cite{platonov2023critical} are considered as heterophilic graphs. The details of datasets are in Appendix \ref{section:appendix_dataset} and Table \ref{tab:datasets}. 

\noindent\textbf{Baselines}. The baselines fall into three categories: (i) Traditional GNN models: GCN \cite{kipf2016semi}, GAT \cite{velivckovic2017graph}, BernNet \cite{he2021bernnet}, ChebNetII \cite{he2022convolutional} and TFE-GNN \cite{duanunifying}; (ii) Graph ``pre-training and fine-tuning'' models: DGI \cite{velivckovic2018deep}, GraphCL \cite{you2020graph}, SimGRACE \cite{xia2022simgrace} and PolyGCL \cite{chen2024polygcl}; (iii) Graph ``pre-training and prompt tuning'' models: GPPT \cite{sun2022gppt}, GPrompt \cite{liu2023graphprompt}, GPF-plus \cite{fang2024universal}, All-in-One \cite{sun2023all}, and ProNoG \cite{yu2024non}. Among them, TFE-GNN, PolyGCL, and ProNoG are designed to handle heterophily. For DGI, GraphCL, SimGRACE, GPPT, GPrompt, GPF-plus, and All-in-One, we use GCN \cite{kipf2016semi} as the backbone model, while other backbones are evaluated in Appendix \ref{section:appendix_graphtransformer}. Details are in Appendix \ref{section:appendix_baseline}.

\noindent\textbf{Implementation Details}. We focus on two scenarios: transductive and inductive settings. In the former, the pre-training graph is the same as the downstream graphs, while in the latter, it's different. Following prior graph prompt work \cite{sun2023all,yu2024non}, we define our task as 5-shot node classification. We set $C$ in the hybrid spectral filter backbone to 2, resulting in 3 different filters. The prompt node number in each prompt graph is 10. Due to label imbalance, we use the F1 score as the metric, with accuracy results in the Appendix \ref{section:appendix_acc}. More details are in Appendix \ref{section:appendix_imple}. Besides the main text, \textbf{we also conduct parameter sensitivity evaluation and runtime analysis in Appendix \ref{section:appendix_parameter} and \ref{section:appendix_runtime}}.

\subsection{Performance Comparison}
\label{section:performance_compar}


Results of transductive learning are in Table \ref{tab:perform_transductive}, inductive in Table \ref{tab:perform_inductive}. More results are in Appendix \ref{section:appendix_acc} and \ref{section:appendix_graphtransformer}.

\noindent\textbf{Transductive Performance}.
From results, we observe:
(i) HS-GPPT achieves state-of-the-art performance, with moderate gains on homophilic graphs and significant improvements on heterophilic ones. Notably, on the \textit{Empire} dataset (homophily 0.047, 18 classes), our method showcases robust performance on this extremely heterophilic dataset.
(ii) Spectral-based baselines (BernNet, ChebNetII, TFE-GNN, PolyGCL) perform well by capturing beyond low-frequency. While some slightly outperform ours on individual datasets (Cornell, Ratings), our method generalizes better via fine-grained spectral alignment.

\noindent\textbf{Inductive Performance} We assess the inductive learning performance across graphs with different homophily levels. We examine both in-domain (where pre-training and downstream graphs are in the same domains) and cross-domain settings (where pre-training and downstream graphs are in different domains). Four dataset pairs are randomly selected, while more are in Appendix \ref{section:appendix_acc} and \ref{section:appendix_graphtransformer}. The results show our model excels in inductive learning, despite the greater challenge of knowledge transfer due to the disparity between pre-trained and downstream graphs. It acquires rich spectral knowledge during pre-training and aligns the downstream graph's spectral distribution with pre-trained knowledge during prompt tuning, enhancing knowledge transfer.

\subsection{Ablation Study}
\label{section:abalation}

We investigate the key modules' significance in pre-training and prompt tuning. In pre-training, we introduce HS-GPPT (low-pass) with a single GCN \cite{kipf2016semi} as GNN backbone. In prompt tuning, we develop three simplified versions: HS-GPPT (single prompt), HS-GPPT (\textit{w/o} prompt), and HS-GPPT (\textit{w/o} prompt norm). HS-GPPT (single prompt) uses one prompt graph for all filters. HS-GPPT (\textit{w/o} prompt) omits prompt graphs, and HS-GPPT (\textit{w/o} prompt norm) skips prompt graph normalization. Partial results are in Table \ref{tab:ablation}, with more in Appendix \ref{section:appendix_ablation}.

\begin{table}
\centering
\scalebox{0.655}{
 \begin{tabular}{c|c|c|c|c|c|c}
 \toprule
  Settings & \multicolumn{3}{c|}{Transductive} & \multicolumn{3}{c}{Inductive} \\ \hline
 Source & Pubmed & Cornell & Squirrel & Chameleon & Pubmed & Squirrel \\
 Target & Pubmed & Cornell & Squirrel & Squirrel & Texas & Cornell  \\  \hline \hline            
                        HS-GPPT & \textbf{0.6910} & \textbf{0.4209} & \textbf{0.2536} & \textbf{0.2307} & \textbf{0.5106}  & \textbf{0.4247}   \\ \hline
                        low-pass  & 0.4294 & 0.2213 & 0.2148 & 0.2143 & 0.3501 & 0.2061  \\ \hline
                        single prompt & 0.6866 & 0.3875 & 0.2285 & 0.2204 & 0.5070 & 0.4203 \\
                        \textit{w/o} prompt & 0.6893 & 0.2869 & 0.2255 & 0.2077 & 0.4121 & 0.3360  \\
                        \textit{w/o} prompt norm & 0.6873 & 0.2793 & 0.2484 & 0.2211 & 0.3011 & 0.3022  \\ \bottomrule
 \end{tabular}}
 \caption{The ablation study of different variants. More results are in Appendix \ref{section:appendix_ablation}.}
\label{tab:ablation}
\vskip -1.5em
\end{table}


Results show: (i) Our method uses pre-trained knowledge and spectral alignment to boost performance and avoid negative transfer. Replacing hybrid filters with a single low-pass filter (HS-GPPT low-pass) hinders alignment, reducing performance. In pre-training, adaptive spectral filter weighting prioritizes the most relevant filters. In downstream tuning, our alignment enables knowledge transfer across graph types, even from homophilic graphs (e.g., PubMed) to downstream heterophilic graphs (e.g., Texas). 
(ii) Finer operations and normalization enhance alignment: HS-GPPT (single prompt) outperforms HS-GPPT (w/o prompt), with the full model best. Normalizing prompt graphs also boosts performance by reducing noise. HS-GPPT (\textit{w/o} prompt norm), without normalization, may introduce noise.

\subsection{Compatibility Investigation}
\label{section:compat_invest}
    
    
\begin{figure}[t]
    \centering
    \includegraphics[width=\columnwidth]{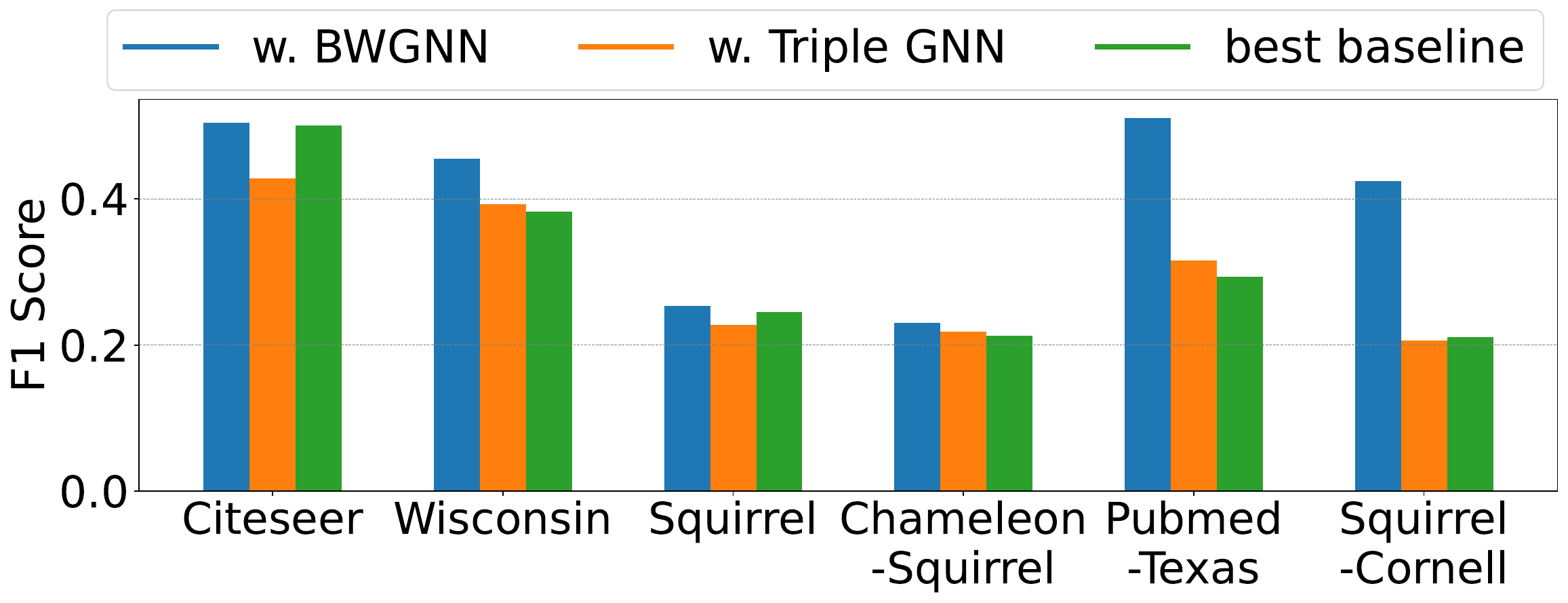} 
    \vskip -0.5em 
    \caption{Compatibility investigation with different hybrid GNN backbones. Here `best baseline' denotes the best baseline results.}
    \vskip -0.8em 
    \label{fig:compatibility}
\end{figure}

\vskip 1em 

\begin{figure}[t]
    \centering
    \includegraphics[width=\columnwidth]{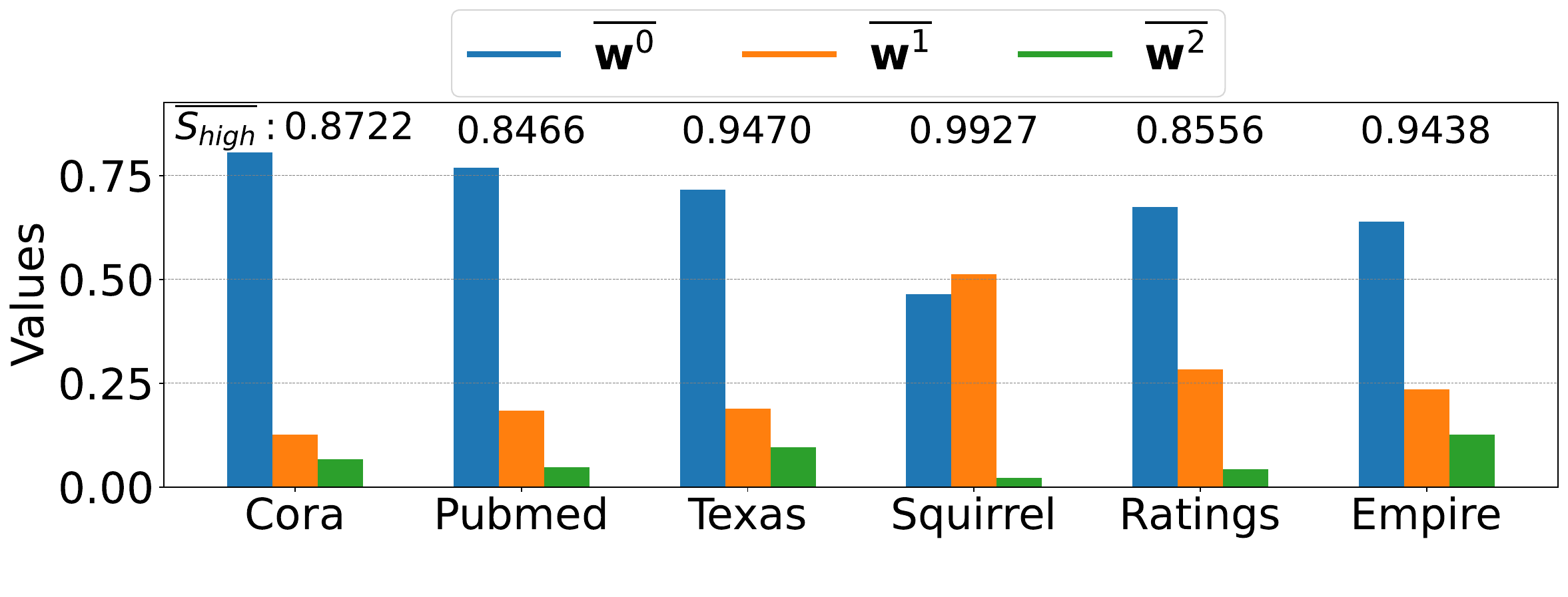}
    \vskip -0.5em
    \caption{Integration weights of filters: $\overline{\mathbf{x}^0}$, $\overline{\mathbf{x}^1}$, $\overline{\mathbf{x}^2}$ correspond to $g_{0,2}$ (low-pass), $g_{1,1}$ (band-pass), and $g_{2,0}$ (low-pass).}
    \vskip -0.8em
    \label{fig:weight_plot}
\end{figure}


We evaluate different GNN backbones by substituting ours with a triple-filter combination (low/mid/high-pass), which we denote as Triple GNN (details are in Appendix \ref{section:appendix_triple_gnn}). Results in Figure \ref{fig:compatibility} show: (i) Abundant spectral knowledge can enhance performance. Compared with the baseline, our model using triple GNN still exhibits competitive performance. (ii) Less spectral overlap among different graph filters can effectively improve performance. BWGNN filters (Figure \ref{figs:bwgnn}) have better spectral locality than Triple GNN (Figure \ref{figs:filter_com}), focusing on narrower bands to act as frequency-specific experts, yielding superior performance.

\subsection{Case Study}

We assess filter significance by computing the average of all dimensions to get $\overline{\mathbf{x}^k}$. Results in Figure \ref{fig:weight_plot} depicts filter weights and the corresponding average $\overline{S_{high}}$ for each datasets. Our analysis uncovers two key insights: (i) Significant filters are prioritized. For instance, low-pass filters dominate in homophilic graphs. (ii) Abundant pre-trained knowledge is necessary. Even in homophilic graphs, non-low-pass filters account for over 10\%, showing the spectral complexity of real-world graphs.

\section{Conclusion}


In this paper, we identify two fundamental limitations in existing graph prompt tuning methods: the knowledge bottleneck caused by single-filter backbones, and the utilization bottleneck arising from the mismatch between pre-trained spectral knowledge and downstream spectra. To overcome this, we propose HS-GPPT, establishing an abundantt spectral basis via hybrid pre-training. Crucially, we treat prompt tuning as a spectral alignment process, employing learnable prompts to align with spectral pre-trained knowledge for fully utilization and better transfer. Experiments show HS-GPPT's superior performance over baselines across different homophily levels in both transductive and inductive settings.

\bibliographystyle{named}
\bibliography{ijcai26}
\clearpage

\clearpage
\appendix
\section{Additional Proofs}

\subsection{Proof of Proposition \ref{prop:spectrum_dis}}
\label{section:appendix_spectrum_dis}

\textit{Proof}. 
Given a graph $\mathcal{G}$ with node label $\mathbf{Y}$ and homophily level $h$, we denote the expectation of normalized intra-class and inter-class distances \cite{ju2025cluster} as:
\begin{gather} \small
    \mathbb{E}(d_{intra}) = \mathbb{E}\left[\frac{(x_u - x_v)^2}{\mathbf{x}^T\mathbf{x}}\right]_{e_{uv}\in\mathcal{E},y_u = y_v} \\
    \mathbb{E}(d_{inter}) = \mathbb{E}\left[\frac{(x_u - x_v)^2}{\mathbf{x}^T\mathbf{x}}\right]_{e_{uv}\in\mathcal{E},y_u \neq y_v},
\end{gather}
where $\mathbb{E}(d_{intra})$ and $\mathbb{E}(d_{inter})$ reflect the node feature distribution. In general, the distance between nodes of different classes is larger than that within the same class, implying $\mathbb{E}(d_{inter}) > \mathbb{E}(d_{intra})$. 

Let $|\mathcal{E}|$ denote the total number of edges. Consequently, the number of homophilic edges is $h|\mathcal{E}|$ and the number of heterophilic edges is $(1-h)|\mathcal{E}|$. By expanding the expectations, we have:
\begin{equation}
    \begin{aligned}
    \mathbb{E}(d_{intra})
    & = \mathbb{E}\left[\frac{(x_u - x_v)^2}{\mathbf{x}^T\mathbf{x}}\right]_{(u,v)\in\mathcal{E},y_u = y_v} \\
    & = {\frac{\sum_{{(u,v)} \in\mathcal{E},y_u=y_v}{(x_u-x_v)}^2}{h|\mathcal{E}|\cdot\mathbf{x}^T\mathbf{x}}},
    \label{equation:d_intra}
    \end{aligned}
\end{equation}
\begin{equation}
    \begin{aligned}
    \label{equation:d_inter}
    \mathbb{E}(d_{inter})
    = \mathbb{E}\left[\frac{(x_u - x_v)^2}{\mathbf{x}^T\mathbf{x}}\right]_{(u,v)\in\mathcal{E},y_u \neq y_v} \\
    = {\frac{\sum_{{(u,v)} \in\mathcal{E},y_u\neq y_v}{(x_u-x_v)}^2}{(1-h)|\mathcal{E}|\cdot\mathbf{x}^T\mathbf{x}}}.
    \end{aligned}
\end{equation}
In spectral graph theory \cite{spielman2007spectral,tang2022rethinking}, using the regular graph Laplacian $\mathbf{L}=\mathbf{D}-\mathbf{A}$, the quadratic form is $\mathbf{x}^T\mathbf{L}\mathbf{x}=\mathbf{x}^T\mathbf{D}\mathbf{x}-\mathbf{x}^T\mathbf{A}\mathbf{x}=\sum_{{(u,v)} \in\mathcal{E}}{(x_u-x_v)}^2$. The high-frequency energy $S_{high}$ is derived as:
\begin{equation}
    \begin{aligned}
    S_{high}
    & =\frac{\mathbf{x}^T\mathbf{L}\mathbf{x}}{\mathbf{x}^T\mathbf{x}} \\
    & =\frac{\sum_{{(u,v)} \in\mathcal{E}}{(x_u-x_v)}^2}{\mathbf{x}^T\mathbf{x}} \\
    &=\frac{\sum_{{(u,v)} \in\mathcal{E},y_u=y_v}{(x_u-x_v)}^2}{\mathbf{x}^T\mathbf{x}} + \\
    & \quad \frac{\sum_{{(u,v)} \in\mathcal{E},y_u \neq y_v}{(x_u-x_v)}^2}{\mathbf{x}^T\mathbf{x}}.
    \label{equation:S_high2}
    \end{aligned}
\end{equation}

With Equation \ref{equation:d_intra} and Equation \ref{equation:d_inter}, we derive Equation \ref{equation:S_high2} as:
\begin{equation}
\begin{aligned}
    S_{high}
    & =h|\mathcal{E}|\cdot\mathbb{E}(d_{intra})+(1-h)|\mathcal{E}|\cdot\mathbb{E}(d_{inter}) \\
    & =|\mathcal{E}|\cdot\Big\{h\cdot\mathbb{E}(d_{intra})+(1-h)\cdot\mathbb{E}(d_{inter})\Big\} \\
    & ={|\mathcal{E}|}\cdot\Big\{\mathbb{E}(d_{inter})-\big[\mathbb{E}(d_{inter})-\mathbb{E}(d_{intra})\big]\cdot h\Big\}
\end{aligned}
\end{equation}

With $\mathcal{C}_{base} = {|\mathcal{E}|}\mathbb{E}(d_{inter})$ and $\mathcal{C}_{gap} = {|\mathcal{E}|}[\mathbb{E}(d_{inter})-\mathbb{E}(d_{intra})]$, we have:
\begin{equation}
    S_{high} = \mathcal{C}_{base} - \mathcal{C}_{gap} \cdot h.
\end{equation}

Since $\mathbb{E}(d_{inter}) > \mathbb{E}(d_{intra})$ and $|\mathcal{E}| > 0$, we have $\mathcal{C}_{gap} > 0$. Therefore, $S_{high}$ is monotonically decreasing with $h$.

\subsection{Proof of Theorem \ref{theorem:spe_spe}}
\label{section:appendix_of_spe_spe}

\textit{Proof}. 
First, we formally introduce the optimization objective. Following \cite{chen2022does}, we focus on binary node classification, as multi-class problems can be decomposed into multiple binary tasks. 
Let $\mathbf{U} \in \mathbb{R}^{N \times N}$ be the eigenvector matrix of the graph Laplacian.
Let $\mathbf{Y} \in \mathbb{R}^{N \times 2}$ be the label matrix, with columns $\mathbf{y}_0, \mathbf{y}_1$ as class indicators. and define the label difference vector $\Delta\mathbf{y} = \mathbf{y}_0 - \mathbf{y}_1$. 
The spectral label is denoted as $\hat{\mathbf{y}}  = \mathbf{U}^T \Delta\mathbf{y} = (\hat{y}_1, \cdots, \hat{y}_N)$, and the spectral signal of node features is $\hat{\mathbf{x}} = \mathbf{U}^T \mathbf{x} = (\hat{x}_1, \cdots, \hat{x}_N)$. 

We employ the Spectral Regression Loss (SRL) \cite{lei2022evennet} to evaluate the quality of the graph filter $g$. The loss is defined as:
\begin{equation}
    \mathcal{L}(\mathcal{G})=\sum_{i=1}^{N}\left(\frac{\hat{y}_i}{\sqrt{N}}-\frac{g(\lambda_i)\hat{x}_i}{\sqrt{\sum_{j=1}^{N}\big({g(\lambda_j)}^2\hat{x}_j^2\big)}}\right)^2.
\end{equation}
A smaller SRL indicates better alignment between the filtered signal and the ground truth labels, implying better downstream performance.

Based on the derivation in previous work~\cite{chen2024polygcl}, for a normalized signal where $\sum_{i=1}^N\hat{x}_i^2=N$, the SRL is bounded by:
\begin{equation}
    \mathcal{L}(\mathcal{G})
    \;\le\;
    2-\frac{2}{cN}\sum_{i=1}^N\hat{x}_i^2\,g(\lambda_i),
\end{equation}
where $c$ is a constant representing the upper bound of $g(\lambda_i)$.
Thus, the upper bound of $\mathcal{L}(\mathcal{G})$ is:
\begin{equation}
    \mathcal{L}_{UB}(\mathcal{G})= 2-\frac{2}{cN}\Big(\sum_{i=1}^N\hat{x}_i^2g(\lambda_i)\Big).
\end{equation}
To minimize the loss upper bound $\mathcal{L}_{UB}(\mathcal{G})$, we must maximize the subtraction term. Discarding constants, the optimization problem becomes:
\begin{equation}
    \max\;\sum_{i=1}^N\hat{x}_i^2\,g(\lambda_i)
    \quad\text{s.t.}\quad
    \sum_{i=1}^N\hat{x}_i^2 = N,\quad \hat{x}_i^2\ge0.
\end{equation}

Let \(w_i = \hat{x}_i^2/N\). Since \(\sum_i w_i=1\) and \(w_i \ge 0\), \(\mathbf{w}\) lies on the probability simplex. Let \(g_i = g(\lambda_i)\). The objective simplifies to maximizing the linear functional \(J = \sum_i w_i g_i\).

Suppose, for the sake of contradiction, that in an optimal weight distribution \(\mathbf{w}^*\), the signal energy is not aligned with the filter response. specifically, there exist indices \(i, j\) such that:
\begin{equation}
    g_i > g_j
    \quad\text{but}\quad
    w^*_i < w^*_j.
\end{equation}
Choose a perturbation \(0 < \varepsilon \le w^*_j\) and define a new weight vector \(\mathbf{w}'\):
\begin{equation}
    w'_i = w^*_i + \varepsilon,
    \quad
    w'_j = w^*_j - \varepsilon,
    \quad
    w'_k = w^*_k
    \quad (k\notin\{i,j\}).
\end{equation}
The vector \(\mathbf{w}'\) remains feasible. Comparing the objectives:
\begin{equation}
    \begin{aligned}
    \sum_{k=1}^N w'_k\,g_k
    &= \sum_{k\neq i,j} w^*_k\,g_k
      + (w^*_i + \varepsilon)\,g_i
      + (w^*_j - \varepsilon)\,g_j \\
    &= \sum_{k=1}^N w^*_k\,g_k \;+\;\varepsilon\,(g_i - g_j).
    \end{aligned}
\end{equation}
Since \(g_i > g_j\) and \(\varepsilon > 0\), we have \(\varepsilon(g_i - g_j) > 0\), which implies:
\begin{equation}
    \sum_{k=1}^N w'_k\,g_k > \sum_{k=1}^N w^*_k\,g_k.
\end{equation}
This contradicts the assumption that \(\mathbf{w}^*\) is optimal. Therefore, to maximize the objective (and minimize SRL), we must have:
\begin{equation}
    g_i > g_j
    \quad\Longrightarrow\quad
    w^*_i \ge w^*_j.
\end{equation}
Recalling \(w_i \propto \hat{x}_i^2\), this proves that larger filter responses \(g(\lambda_i)\) must be paired with larger signal energy \(\hat{x}_i^2\) (i.e., positive correlation) to minimize the loss.

\subsection{Proof of Theorem \ref{theorem:spec_ada}}
\label{appendix:spec_ada}
\textit{Proof}. For a graph $\mathcal{G}=(\mathcal{V},\mathbf{X},\mathcal{E})$ with its adjacent matrix $\mathbf{A}$, since $\mathcal{V}$ is related to $\mathbf{X}$ and $\mathcal{E}$ is related to $\mathbf{A}$, the graph can also be represented as $\mathcal{G}=(\mathbf{A},\mathbf{X})$. With the notations, to illustrate Theorem \ref{theorem:spec_ada}, we first propose Proposition \ref{pro:spectral_uni}.
\begin{proposition}
    \label{pro:spectral_uni}
    Given graph $\mathcal{G}_1={(\mathbf{A}_1,\mathbf{X}_1})$ and graph $\mathcal{G}_2=(\mathbf{A}_2,\mathbf{X}_2)$ with distinct spectral distributions, i.e., $S_{high1} \neq S_{high2}$. 
    Let $t$ denote any graph-level transformation like ``changing node features'' or ``adding or removing edges/subgraphs'', represented as $t:\mathbb{G}\xrightarrow{}\mathbb{G}$. Then the transformed graph is $\hat{\mathcal{G}}=(\mathbf{\hat{A}},\mathbf{\hat{X}})=t(\mathbf{A},\mathbf{X})$. There exists a graph-level transformation $t^{*}$ such that:
    \begin{equation}
        \mathcal{G}_2=(\mathbf{A}_2,\mathbf{X}_2)=t^{*}(\mathbf{A_1},\mathbf{X}_1).
    \end{equation}
\end{proposition}
\textit{Proof.} Since $S_{high}=\frac{\mathbf{x}^T\mathbf{L}\mathbf{x}}{\mathbf{x}^T\mathbf{x}}$, for graphs $\mathcal{G}_1$ and $\mathcal{G}_2$ where $S_{high1} \neq S_{high2}$, we have:
\begin{equation}
    \frac{\mathbf{x}_1^T\mathbf{L}_1\mathbf{x}_1}{\mathbf{x}_1^T\mathbf{x}_1} \neq \frac{\mathbf{x}_2^T\mathbf{L}_2\mathbf{x}_2}{\mathbf{x}_2^T\mathbf{x}_2}.
\end{equation}
Since here $\mathbf{L}=\mathbf{D}-\mathbf{A}$, the above inequation can be written as:
\begin{equation}
    \frac{\mathbf{x}_1^T(\mathbf{D}_1-\mathbf{A}_1)\mathbf{x}_1}{\mathbf{x}_1^T\mathbf{x}_1} \neq \frac{\mathbf{x}_2^T(\mathbf{D}_2-\mathbf{A}_2)\mathbf{x}_2}{\mathbf{x}_2^T\mathbf{x}_2}.
    \label{equ:tem_1}
\end{equation}
Here $\mathbf{D}$ is related to $\mathbf{A}$. Therefore, when Inequality \ref{equ:tem_1} is satisfied, the following three scenarios exist:
\begin{itemize}
    \item $\mathbf{X}_1\neq\mathbf{X}_2, \mathbf{A}_1=\mathbf{A_2}$. In such case, ${\mathcal{G}_2}$ has the same adjacency matrix as ${\mathcal{G}_1}$, while only the node features differ. Therefore, by modifying the node features ${\mathbf{X}_1}$, we can transform ${\mathcal{G}_1}$ into ${\mathcal{G}_2}$. 
    \item $\mathbf{X}_1=\mathbf{X}_2, \mathbf{A}_1\neq\mathbf{A}_2$. In such case, $\mathcal{G}_2$ can be obtained by modifying the edges $\mathbf{A}_1$. 
    In this scenario, ${\mathcal{G}_2}$ has the same node features as ${\mathcal{G}_1}$, while only the adjacency matrix differ. Therefore, by adding or removing edges, we can transform ${\mathcal{G}_1}$ into ${\mathcal{G}_2}$. 
    \item $\mathbf{X}_1\neq\mathbf{X}_2, \mathbf{A}_1\neq\mathbf{A}_2$. In such case, both the adjacent matrix and node features of $\mathcal{G}_1$ and $\mathcal{G}_2$ are different. There, we can transform $\mathcal{G}_1$ into $\mathcal{G}_2$ through a combination of transformation operations (e.g., ``changing node features'', ``adding or deleting edges'', and ``adding or removing isolated sub-graphs'') \cite{fang2024universal}.
\end{itemize}
Therefore, by performing graph-level transformation operations on $\mathcal{G}_1$, we can obtain $\mathcal{G}_2={(\mathbf{A}_2,\mathbf{X}_2})$.

Proposition \ref{pro:spectral_uni} indicates that through performing graph-level transformations on a graph, its spectral distribution can be changed into an arbitrary distribution.

Then we introduce Lemma \ref{lemma:tem_1} from \cite{sun2023all}:
\begin{lemma}
    \label{lemma:tem_1}
    Given a pre-trained model $F$, and an input graph $\mathcal{G}=(\mathbf{A},\mathbf{X})$. Let $t$ be any graph-level transformation. There exists a prompt graph $G_p$ that satisfies:
    \begin{equation}
        F\Big(\psi(\mathcal{G},\mathcal{G}_p)\Big)=F\Big(t(\mathbf{A},\mathbf{X})\Big).
    \end{equation}
\end{lemma}
Lemma \ref{lemma:tem_1} demonstrates that the prompt graph can simulate arbitrary graph-level transformations, ensuring that the output of the model is approximately equal to the graph after any graph-level transformation.

With Lemma \ref{lemma:tem_1} and Proposition \ref{pro:spectral_uni}, we have:
\begin{equation}
    F\Big(\psi(\mathcal{G}_1,\mathcal{G}_p)\Big)=F\Big(t^{\star}(\mathbf{A_1},\mathbf{X_1})\Big)=F(\mathbf{A}_2,\mathbf{X}_2)=F(\mathcal{G}_2).
\end{equation}
Therefore, when graphs are fed into the pre-trained model, the prompt graph is capable of arbitrarily modifying the spectral distribution of the input graph. This arbitrary transformation ability enables the spectral distribution of downstream graphs to align with that of pretexts, thus benefiting the transfer of pre-trained knowledge.

\subsection{Proof of Corollary \ref{corollary:srl_bound}}
\label{appendix:corollary}
\textit{Proof}. The pre-trained graph filter is $g(\mathbf{\Lambda})$, in the proof of Theorem \ref{theorem:spe_spe} (Appendix \ref{section:appendix_of_spe_spe}), we show
\begin{equation}
    \mathcal{L}_{UB}(\mathcal{G}) = 2-\frac{2}{cN}\Big(\sum_{i=1}^N\hat{x}_i^2g(\lambda_i)\Big).
\end{equation}
We denote the prompted graph $\tilde{\mathcal{G}}=\psi(\mathcal{G},\mathcal{G}_p)$, thus the original spectral graph signal is $\hat{\mathbf{x}}$ and the prompted spectral graph signal is $\hat{\tilde{\mathbf{x}}}$. 

Following prior work \cite{fang2024universal,sun2023all}, the effect of introducing prompt nodes can be equivalently viewed as a transformation on the features of the original nodes. Since our analysis ultimately focuses on the performance on these original nodes, we therefore maintain a fixed node number in our theoretical framework, modeling the prompt's effect as a change from the original signal $\hat{\mathbf{x}}$ to a new signal $\hat{\tilde{\mathbf{x}}}$. Therefore, the SRL of the two graphs are:
\begin{gather}
    \label{equation:l_ub1}
    \mathcal{L}_{UB}(\mathcal{G}) = 2-\frac{2}{cN}\Big(\sum_{i=1}^N\hat{x}_i^2g(\lambda_i)\Big) \\
    \label{equation:l_ub2}
    \mathcal{L}_{UB}(\psi(\mathcal{G},\mathcal{G}_p)) = 2-\frac{2}{cN}\Big(\sum_{i=1}^N\hat{\tilde{x}}_i^2g(\lambda_i)\Big)
\end{gather}
In Theorem \ref{theorem:spec_ada}, we prove that the prompt graph can simulate any spectral distribution. Therefore, there exists a prompt graph $\mathcal{G}_p$ such that
\begin{equation}
    \label{equation:ub}
    \sum_{i=1}^N\hat{\tilde{x}}_i^2g(\lambda_i) \geq\sum_{i=1}^N\hat{x}_i^2g(\lambda_i).
\end{equation}
This inequality demonstrate the prompt graph can align the spectral distribution of downstream graphs with pretexts. 

Combine Equation \ref{equation:l_ub1}, \ref{equation:l_ub2} and Inequality \ref{equation:ub}, we have:
\begin{equation}
       \mathcal{L}_{UB}(\psi(\mathcal{G},\mathcal{G}_p)) \leq \mathcal{L}_{UB}(\mathcal{G}) 
\end{equation}

\section{Experiment Setup Details}

\subsection{Baselines Descriptions}
\label{section:appendix_baseline}
To verify the effectiveness of our proposed model, we compare it with various models, which can be divided into three groups: traditional GNN models, graph ``pre-training and fine-tuning'' models, and graph ``pre-training and prompt tuning'' models. 

The first group includes traditional GNN models. They are trained using supervised learning on the graph and then perform inference on the same graph.
\begin{itemize}
    \item \textbf{GCN} \cite{kipf2016semi}: This method achieves neighborhood information aggregation by spectral graph convolutions.
    \item \textbf{GAT} \cite{velivckovic2017graph}: This method proposes an attention mechanism to aggregate neighborhood information.
    \item  \textbf{BernNet} \cite{he2021bernnet}: This spectral method overcomes the limitations of oversimplified or ill-posed filters. It estimates any filter over a graph's normalized Laplacian spectrum via an order-K Bernstein polynomial approximation and designs spectral properties by setting Bernstein basis coefficients.
    \item  \textbf{ChebNetII} \cite{he2022convolutional}: This spectral method is based on Chebyshev interpolation, which enhances the original Chebyshev polynomial approximation while reducing the Runge phenomenon.
    \item  \textbf{TFE-GNN} \cite{duanunifying}: This spectral method generalizes both homophily and heterophily through ensembles of multiple spectral filters.
\end{itemize}
The second group includes graph ``pre-training and fine-tuning'' models. They propose self-supervised tasks in the pre-training stage and then fine-tune the model on downstream graphs.
\begin{itemize}
    \item \textbf{DGI} \cite{velivckovic2018deep}: DGI functions as a self-supervised pre-training method designed for homogeneous graphs. It is based on the maximization of mutual information (MI), with the aim of increasing the estimated MI between locally augmented instances and their global equivalents.
    \item \textbf{GraphCL} \cite{you2020graph}: GraphCL utilizes a range of graph augmentations for self-supervised learning, exploiting the inherent structural patterns of graphs. The main objective is to enhance the consistency between different augmentations during graph pre-training.
    \item  \textbf{SimGRACE} \cite{xia2022simgrace}: SimGRACE overcomes the limitations of existing GCL methods related to data augmentations. It utilizes the original graph and a perturbed GNN model as encoders to generate contrastive views without the need for data augmentations. Additionally, it employs an adversarial training scheme to boost robustness, thereby attaining competitive performance along with high flexibility and efficiency.
    \item  \textbf{PolyGCL} \cite{chen2024polygcl}: PolyGCL is spectral-based and solves the limitations of current methods when dealing with heterophilic graphs. It employs polynomial filters to produce spectral views for contrastive learning between low-pass and high-pass views, incorporating high-pass information. This enables PolyGCL to show superiority on graphs with varying homophily.
\end{itemize}
The third group includes graph ``pre-training and prompt tuning'' models. They reformulate downstream tasks into pretexts, thereby facilitating knowledge transfer.
\begin{itemize}
    \item \textbf{GPPT} \cite{sun2022gppt}: GPPT utilizes the link prediction task to pre-train a GNN model and reformulates the downstream node classification task into the link prediction task. However, it has been proven that the link prediction task only preserves the low-frequency similarity information \cite{liu2022revisiting,yu2024non}.
    \item \textbf{GPrompt} \cite{liu2023graphprompt}: GraphPrompt employs subgraph similarity prediction to pre-train the GNN model and reformulates the downstream node and graph classification into the subgraph similarity prediction template. However, the subgraph similarity prediction task is similar to the link prediction task, which only captures low-pass information.
    \item  \textbf{GPF-plus} \cite{fang2024universal}: GPF proposes a universal prompt-based tuning method for pre-trained GNN models, regardless of the pre-training strategy. It operates within the input graph's feature space and theoretically attains an equivalent effect to any prompting function, thereby eliminating the need for an explicit illustration of prompting functions for each strategy. Here, we utilize the more flexible and robust version, GPF-plus.
    \item  \textbf{All-in-One} \cite{sun2023all}: All-in-One reformulates all the tasks into sub-graph classification and adopts prompt graphs based on GPF-plus \cite{fang2024universal} to adapt universally to any pre-training strategy.
    \item  \textbf{ProNoG} \cite{yu2024non}: ProNoG addresses the limitations of existing graph prompt methods for non-homophilic graphs. ProNoG first analyzes existing graph pre-training methods to provide theoretical insights on pre-training task choices, and then proposes a conditional network to characterize node-specific non-homophilic patterns in downstream tasks.
\end{itemize}

\subsection{Implementation Details}
\label{section:appendix_imple}
To conduct the experiment under inductive learning, we first apply SVD (Singular Value Decomposition) to reduce the initial features to 128 dimensions. For pre-training and prompt tuning, we set the learning rates to $1e^{-3}$ and $5e^{-3}$, respectively. The parameter $C$ in the hybrid graph filter backbone is set to 2, resulting in 3 graph filters. In each prompt graph, the number of prompt nodes is 10. Additionally, we set the inner edge threshold $\tau_{inner}$ to 0.2. For cross edges threshold $\tau_{cross}$ , we set it to 0.55 in homophilic graphs (i.e., Cora, Pubmed, and Citeseer) and 0.4 for other heterophilic graphs.

We define our task as 5-shot learning. Specifically, for each class, we randomly select 5 samples for prompt tuning. We then divide the remaining data into two equal parts, with one part used as the validation set and the other as the test set. We run the experiments 5 times with different random seeds and obtain the average of the results. The model is trained for 2000 epochs, and we evaluate the model based on its best F1 score on the validation set. All the experiments are conducted with torch 1.3.11 on the NVIDIA GeForce RTX 3090 GPU.

For the baseline implementations, including GCN, GAT, DGI, GraphCL, SimGRACE, GPPT, GPrompt, GPF-plus, and All-in-One, we leveraged the public benchmark ProG \cite{zi2024prog} to implement them using both 2-layer GCN and UniMP (a Graph Transformer model) as the backbone GNNs. For BernNet, ChebNetII, TFE-GNN, PolyGCL, and ProNoG, we use the source code published by the original authors. For baselines involving a pre-training stage, we adopted the hyperparameters recommended in the respective source code to pre-train the models. During the downstream tuning phase, hyperparameters such as the learning rate and epochs were kept consistent with those of our proposed methods to ensure a fair comparison.

For homophilic graph prompt baselines implemented via ProG, we follow the ProG benchmark to perform a grid search for optimal pretraining methods on each dataset. Specifically, we evaluate six pretraining methods: DGI \cite{velivckovic2018deep}, GraphMAE \cite{hou2022graphmae}, EdgePreGPPT \cite{sun2022gppt}, EdgePreGprompt \cite{liu2023graphprompt}, GraphCL \cite{you2020graph}, and SimGRACE \cite{xia2022simgrace}.
For the heterophilic graph prompt baseline ProNoG, we adopt the two pretraining methods recommended in its original paper: DGI \cite{velivckovic2018deep} and GraphCL \cite{you2020graph}.
We report the best results from combinations of pretraining methods and prompts for each dataset.
Optimal pretraining-prompt tuning combinations are detailed in Tables \ref{tab:appendix_combination_trans_1}, \ref{tab:appendix_combination_trans_2}, and \ref{tab:appendix_combination_inductive}.

\subsection{Implementation Details}
\label{section:appendix_dataset}
\begin{table*}
\centering
\scalebox{1}{
\setlength{\tabcolsep}{0.75 mm}{
\begin{tabular}{@{}cc|ccccc@{}}
\toprule
& Dataset     & \# Nodes & \# Edges &  \# Features & \# Class & Homophily  \\ \hline \hline
& Cora      & 2708     & 5429   & 1433 & 7  & 0.810    \\ \hline
& Citeseer    &  3327     & 4552  & 3703  & 6 & 0.736    \\ \hline
& Pubmed   &  19717   & 44324  & 500  & 3 & 0.802    \\ \hline
& Texas   &  183   & 325    & 1703 & 5 & 0.108  \\ \hline
& Cornell   &  183   & 298    & 1703 & 5 & 0.305    \\ \hline
& Wisconsin   &  251   & 515    & 1703  & 5 & 0.196   \\ \hline
& Chameleon   &  2277   & 36101    & 2277 & 5  & 0.235   \\ \hline
& Squirrel   &  5201   & 217073    & 2089  & 5 & 0.224 \\ \hline
& Amazon-ratings   &  24492   & 186100    & 300  & 5 & 0.380 \\ \hline
& Roman-empire   &  22622   & 65854    & 300  & 18 & 0.047    \\  \bottomrule
\end{tabular}}}
\caption{Statistics of experiment datasets.}
\label{tab:datasets}
\end{table*}
Here we provide a detailed introduction to the real-world datasets below, and the statistics of the datasets are presented in Table \ref{tab:datasets}.
\begin{itemize}
    \item \textit{Cora} \cite{mccallum2000automating}, \textit{Citeseer} \cite{sen2008collective}, and \textit{Pubmed} \cite{kipf2016semi} are three citation networks that are regarded as typical homophilic graphs. Within these networks, the nodes stand for papers, and the edges signify the citation connections between two papers. The features are comprised of bag-of-word representations of the papers, and the labels serve to denote the research topic of each paper.
    \item \textit{Cornell}, \textit{Texas}, and \textit{Wisconsin} \cite{pei2020geom} are three heterophilic networks that come from the WebKB project. In these networks, the nodes are web pages of computer science departments of different universities, and the edges are hyperlinks between them. The features of each page are represented as bag-of-words, and the labels indicate the types of web pages.
    \item \textit{Chameleon} and \textit{Squirrel} \cite{rozemberczki2021multi} are two heterophilic networks built upon Wikipedia. In these networks, the nodes represent web pages within Wikipedia, and the edges represent the links that connect them. The features are composed of informative nouns found on the Wikipedia pages, and the labels show the average traffic that the web pages receive.
    
    \item \textit{Amazon-ratings} and \textit{Roman-empire} \cite{platonov2023critical} are substantial heterophilic graphs that possess distinct structural properties and originate from different fields. These graphs are proposed to mitigate the issues present in existing heterophilic graphs. For instance, the networks of \textit{Chameleon} and \textit{Squirrel} contain a significant number of duplicate nodes, which results in problems such as training and test data leakage. 
    
    In detail, \textit{Roman-empire} is a word dependency graph based on the Roman Empire article from the English Wikipedia and \textit{Amazon-ratings} is a product co-purchasing network.
\end{itemize}

\subsection{Synthetic Dataset Details}
\label{section:appendix_csbm}
We utilize the widely-employed Contextual Stochastic Block Model (CSBM) \cite{chien2020adaptive} to generate synthetic datasets for validating our theorems. Specifically, these graphs are distinguished by adjustable edge probabilities both within and between different classes. The fundamental concept is that nodes within the same class display a uniform feature distribution. The graph is generated as $\mathcal{G} \sim CSBM(n,f,\sigma,\mu)$, where $n$ stands for the number of nodes, $f$ represents the feature dimension and $\sigma$ and $\mu$ are hyperparameters that affect the graph structure and node features, respectively.

We generate two classes of equal size, namely $c_0$ and $c_1$, each containing $n/2$ nodes. The node features are generated using the following formula:
\begin{equation}
    \label{equation:csbm}
    \mathbf{x}_i=\sqrt{\frac{\mu}{n}}y_iu+\frac{w_i}{\sqrt{f}},
\end{equation}
where $y_i\in \{-1,+1\}$ indicates the label of node $v_i$, $\mu$ is the mean value of the Gaussian distribution $u\sim\mathcal{N}(0,I/f)$, and the elements of $w_i$ follow independent standard normal distributions.

The graph structure is generated as follows:
\begin{equation}
    P(\mathbf{A}_{ij}=1) = \begin{cases} 
\frac{1}{n}(d+\sigma\sqrt{d}) & \text{when } y_i=y_j \\
 \frac{1}{n}(d-\sigma\sqrt{d})& \text{when} y_i\neq y_j.
\end{cases}
\end{equation}
Here $d$ is the average degree of nodes. Following previous research \cite{pmlr-v235-wan24g}, we can adjust the homophily level $h$ by modifying $\sigma=\sqrt{d(2h-1)}$, thereby obtaining graphs with different structures. For our validation experiments, we set $d = 50$, $n = 3000$, and $f = 128$, and generate synthetic graphs with the homophily level $h$ ranging from 0 to 1.

\subsection{Experiments on Synthetic Datasets}
\label{section:appendix_triple_gnn}
\begin{figure*}
    \centering
    \includegraphics[scale=0.29]{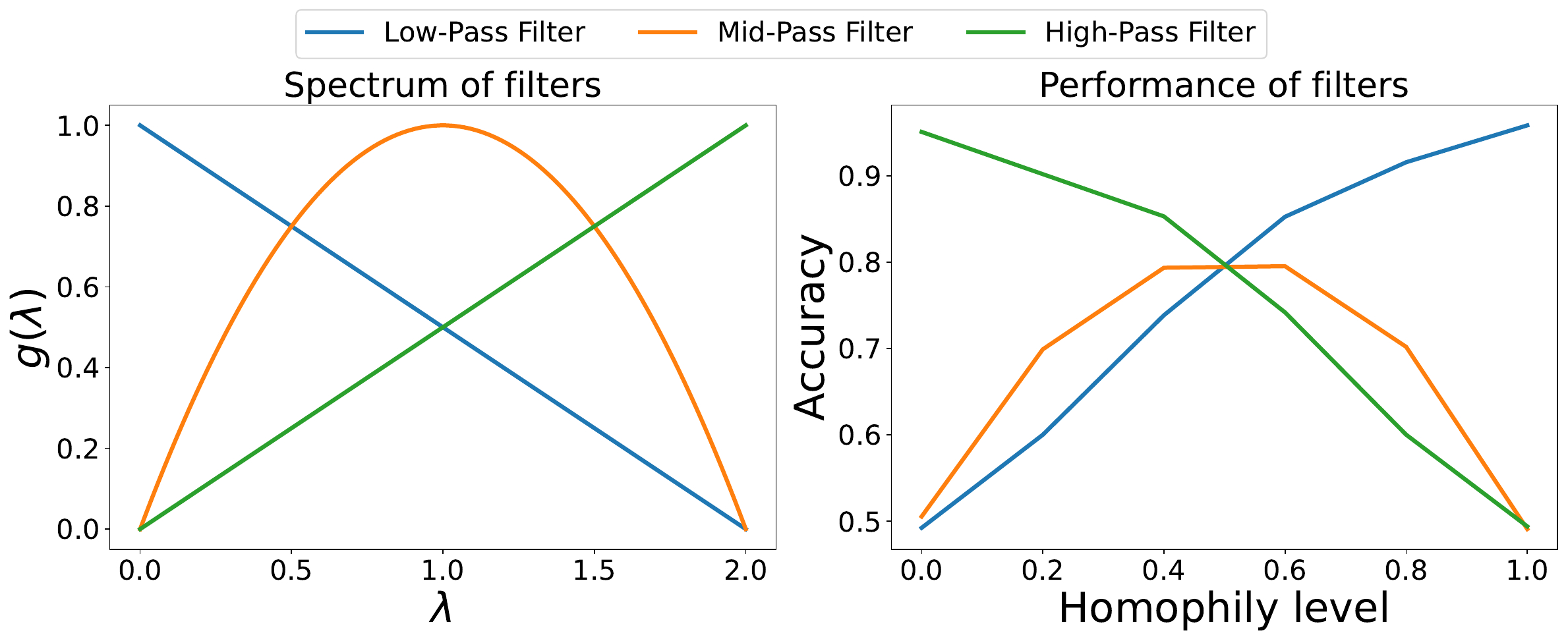}
    \caption{Case studies on graphs with varying homophily levels. The left figure illustrates the spectrum  $g(\mathbf{\Lambda})$  of different filters, while the right figure shows the performance on graphs with homophily levels from 0 to 1.}
    \label{figs:filter_com}
\end{figure*}

To present our theoretical findings in a more intuitive way, we further conduct the validation experiments to verify Theorem \ref{theorem:spe_spe}. We utilize GNNs with different filtering characteristics to learn the CSBM graphs graph generated in Figure \ref{fig:s_high_violin}. To avoid complex calculations, following previous work \cite{luo2024spectral}, we use functions with linear or quadratic terms to construct the graph filter. We adopt three graph filters: low-pass, mid-pass, and high-pass filters, which can be formulated as:
\begin{align}
   & g_{low}(\mathbf{\Lambda})=1-\lambda/2 \\
   & g_{mid}(\mathbf{\Lambda})=-(\lambda-1)^2+1 \\
   & g_{high}(\mathbf{\Lambda})=\lambda/2.
\end{align}
We apply local-global contrastive learning \cite{hassani2020contrastive} and subsequently evaluate the performance through a node classification task. To present the performance differences of different graph filters more clearly, we set the ratio of the training set, validation set, and test set to 5:2:3. The visualization and results of the three filters are shown in Figure \ref{figs:filter_com} and are consistent with Theorem \ref{theorem:spe_spe}, indicating that a single graph filter cannot generalize across graphs with varying spectral distributions.

Specifically, low-pass GNNs can show better performance when dealing with graph-structured data that has a relatively high proportion of low-frequency components. Similarly, mid-pass GNNs can highlight their performance advantages on graphs where mid-frequency components are predominant. And for graphs rich in high-frequency components, high-pass GNNs can achieve the best performance.

\section{Methodology Details}
\subsection{Beta Wavelet GNNs}
\label{section:appendix_bwgnn}
\begin{figure*}
    \centering
    \includegraphics[scale=0.30]{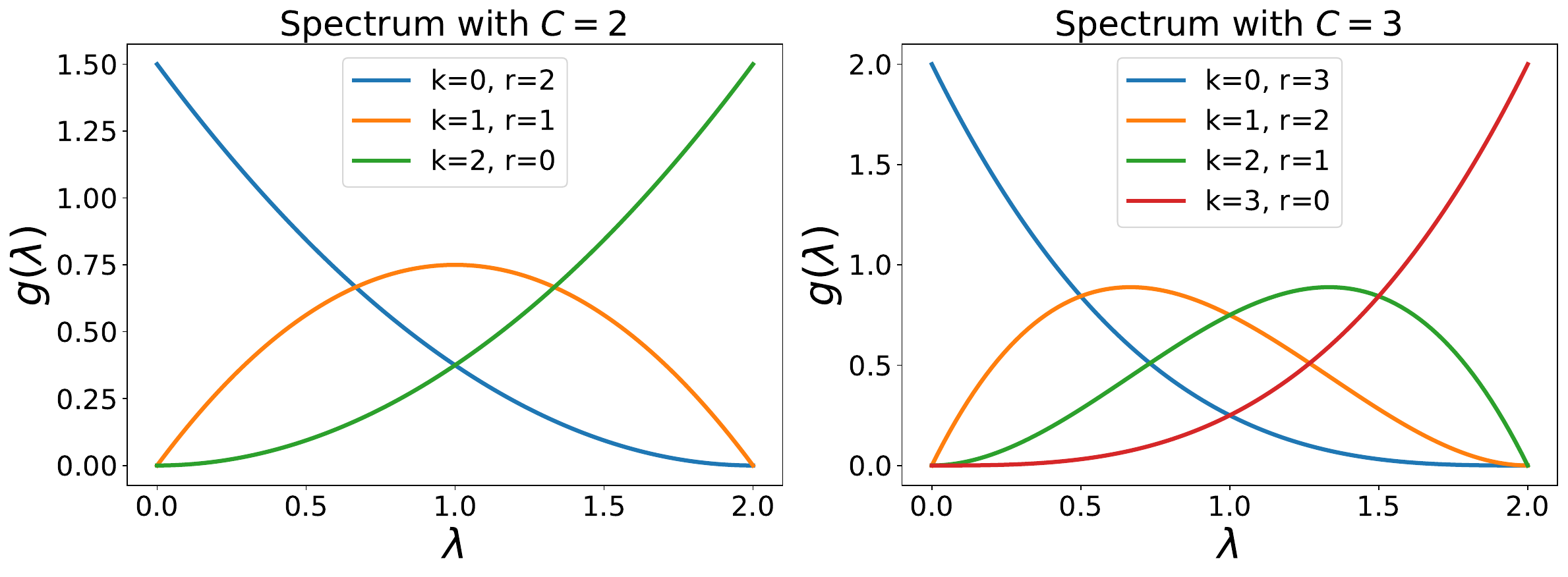}
    \caption{Spectrum of Beta Wavelet GNNs with $C=2$ and $C=3$. It is constructed by a group of different filters.}
    \label{figs:bwgnn}
\end{figure*}

The probability density of the Beta distribution admits:
\begin{equation}
    g_{k,r}(w)= \begin{cases}
    \frac{1}{B(k+1,r+1)}w^k{(1-w)}^r, & if \ w \in [0,1] \\
    0, & otherwise
    \end{cases}
\end{equation}
where $k,r \in \mathbb{R}^+$ and $B(k+1,r+1)=\frac{k!r!}{(k+r+1)!}$ is a constant. Since the eigenvalues of the normalized graph Laplacian $L$ satisfy $\lambda \in [0,2]$, we utilize $g_{k,r}^*(w)=\frac{1}{2}g_{k,r}\frac{w}{2}$ to convert all the frequencies. To restrict $k,r \in \mathbb{N}^+$, the graph filter based on Beta wavelet transformation is:
\begin{equation}
    g_{k,r}(\mathbf{L})=\frac{{\frac{\mathbf{L}}{2}}^k{(\mathbf{I}-\frac{\mathbf{L}}{2})}^r}{2B(k+1,r+1)}.
\end{equation}
With different values of $k$ and $r$, $g_{k,r}$ has different filtering characteristics. When we set $k+r=C$ as a constant, we can obtain a group of $C+1$ graph filters with different spectral characteristics. 
\begin{equation}
    g_{\beta}=(g_{0,C},g_{1,C-1},\cdots,g_{C,0})
\end{equation}
Here $g_{k,r}$ is a $C$ power polynomial which denotes $C$-hop neighbors are considered. In this way, we obtain a group of filters with different spectral characteristics. Specifically, $g_{0,C}$ is a low-pass filter, $g_{C,0}$ is a high-pass filter, and others are band-pass filters of different scales. We visualize the spectral characteristics with $C=2$ and $C=3$ in Figure \ref{figs:bwgnn}. 

\subsection{Algorithm of HS-GPPT}
\begin{algorithm}[t]
\caption{HS-GPPT Pre-Training Framework}
\label{alg:pretraining_algorithm}
\textbf{Input}: A pre-training graph $\mathcal{G} = {\{\mathcal{V}, \mathcal{\mathbf{X}}, {\mathcal{E}}}\}$.\\
\textbf{Parameter}: The number of training epochs $N_{epoch}$, the embedding size $d$, and the $C$ order.\\
\textbf{Output}: The pre-trained graph filters $g_{k,C-k}$ and its corresponding integration weights $\mathbf{w}^k$.
\begin{algorithmic}[1]
\FOR{$t = 1, ..., N_{epoch}$}
    \STATE Obtain $\mathbf{x}_i^{-}$ by shuffling.
    \STATE \emph{// obtain representations from each graph filter}
    \FOR{$k=0,...,C$}
        \STATE Obtain $\mathbf{z}_i^k$ via Equation \ref{equation:tem_1}.
    \ENDFOR
    \STATE \emph{// linear integration from different graph filters}
    \STATE Obtain $\mathbf{z}_i$ via Equation \ref{equation:pre_inte}.
    \STATE \emph{// mean pooling}
    \STATE Obtain $\mathbf{z_{g}} =\frac{1}{N}\sum_{i=1}^N\mathbf{z}_i$.
    \STATE \emph{// obtain the negative representations}
    \FOR{$k=0,...,C$}
        \STATE Obtain $\mathbf{z}_i^{k-}$ via Equation \ref{equation:tem_1}.
    \ENDFOR
    \STATE Minimize loss via Equation \ref{equation:pre_loss}.
\ENDFOR
\end{algorithmic}
\end{algorithm}

\begin{algorithm}[t]
\caption{HS-GPPT Prompt Tuning Framework}
\label{alg:prompt_algorithm}
\textbf{Input}: A downstream graph $\mathcal{G} = {\{\mathcal{V}, \mathcal{\mathbf{X}}, {\mathcal{E}}}\}$, A group of frozen pre-trained graph filter $g_{k,C-k}$ and its corresponding integration weights $\mathbf{w}^k$\\
\textbf{Parameter}: The number of training epochs $N_{epoch}$, the embedding size $d$, the nodes number $N_p$ in each prompt graph and the $C$ order.\\
\textbf{Output}: Optimized parameters of prompt nodes $\{\mathbf{p}_i^k,...,\mathbf{p}_{N_p}^k\}$ of each prompt graph $\mathcal{G}_p^k$.
\begin{algorithmic}[1] 
\FOR{$t = 1, ..., N_{epoch}$}
    \STATE \emph{// obtain representations from each graph filter}
    \FOR{$k=0,...,C$}
        \STATE Normalize prompt nodes $\mathbf{p}^k_i$ via Equation \ref{equation:prompt_norm}.
        \STATE Construct inner edges in $G_p^k$ via $e_{ij} = \mathbb{I}\Big({\sigma(\mathbf{p}_i^k\cdot {\mathbf{p}_j^k}^T) > \tau_{inner}}\Big)$.
        \STATE Obtain $\tilde{\mathbf{x}}_i^k$ by inserting the prompt graph $G_p^k$ via $e_{ij} = \mathbb{I}\Big({\sigma({\mathbf{p}_i^k}^{\prime}\cdot \mathbf{x}_j^T) > \tau_{cross}}\Big)$.
        \STATE Obtain $\tilde{\mathbf{z}}_i^k$ via Equation \ref{equation:prompt_agg_inte}.
    \ENDFOR
    \STATE \emph{// linear integration from different graph filters}
    \STATE Obtain $\tilde{\mathbf{z}}_i$ via Equation \ref{equation:prompt_agg_inte}.
    \STATE Minimize loss via Equation \ref{equation:prompt_loss}.
\ENDFOR
\end{algorithmic}
\end{algorithm}

The algorithm of HS-GPPT is shown in Algorithm \ref{alg:pretraining_algorithm} and \ref{alg:prompt_algorithm}. In the pre-training stage, we train the graph filters and the integration weights. In the prompt tuning stage, we keep the graph filters and integration weights frozen and only tune the learnable prompt graphs and task head (i.e., one-layer MLP).

\subsection{Complexity Analysis}
\label{section_appendix_complexity}
Given an input graph consisting of $N$ nodes and $|\mathcal{E}|$ edges, we assume that each prompt graph $G_p$ contains $N_p$ nodes and $\mathcal{E}_p$ edges. We denote the number of graph filters as $C'=C+1$, where $C$ is the polynomial order. 

For a $C'$-layer GCN \cite{kipf2016semi}, the parameter complexity is $O(C'd^2)$. In our pre-training stage, the tuned parameter is $O(C'd^2 + C'd)$. Due to the fact that $C' \ll d$, our pre-trained parameters are lightweight. In the prompt-tuning stage, we freeze the pre-trained graph filters and integration weights, so the tuned parameter is $O(C'N_pd)$. Since both $C' \ll d$ and $N_p \ll d$, our prompt tuning is also lightweight.

Regarding the time complexity, a $C'$-layer GCN requires $O(C'Nd^2 + C'|\mathcal{E}|d)$ to complete the propagation process and generate node representations. In contrast, our model has a time complexity of $O\Big(C'(N + N_p)d^2 + C'(|\mathcal{E}| + |\mathcal{E}_p|)d + C'(N + N_p)d\Big)$ during prompt tuning. Compared with the original time, the additional time we introduce is $O\Big(C'N_pd^2 + C'(|\mathcal{E}_p|)d + C'(N + N_p)d\Big)$. Given that $N_p \ll d$, $N_p \ll N$, and $M_p \ll M$, we only introduce a very limited amount of additional time. We also carry out the runtime experiment in Appendix \ref{section:appendix_runtime}, which shows the relatively short runtime of our model.

\section{Related Work}
\label{appendix:related_work}
\subsection{Graph prompt tuning}

Graph Neural Networks (GNNs) \cite{kipf2016semi,defferrard2016convolutional,velivckovic2017graph} have emerged as a powerful framework for learning from graph data, which enable information propagation and feature extraction through iterative message passing among neighboring nodes. Typically, GNNs operate under a supervised setting, where models are trained for a specific task on the input graph and make inferences on the same graph. However, the difficulty of obtaining labeled data \cite{chen2024llaga} limits the performance. To break through this limitation, some research \cite{jin2020self,you2020graph,xia2022simgrace,zhu2021graph,hou2023graphmae2,xia2023automated} have turned to self-supervised learning on graph data. 
These methods first pre-train a graph model using self-supervised tasks, and then fine-tune it with downstream tasks. This approach enables the pre-trained knowledge to enhance the performance of the downstream tasks. Nevertheless, it is proven that the misalignment between the pretexts and downstream tasks will impede knowledge transfer and even lead to negative transfer \cite{wang2021afec}.

In response to this challenge, graph prompt tuning \cite{sun2022gppt,fang2024universal,sun2023all,liu2023graphprompt,yu2024non} has emerged as an attractive alternative to the conventional fine-tuning paradigm. The core idea is to design prompts to manipulate the downstream tasks, allowing them to better align with the frozen pre-trained model. For example, GPrompt \cite{liu2023graphprompt} uses prompt vectors to unify pre-training and downstream tasks under a common template, while GPF \cite{fang2024universal} inserts prompt nodes adaptable to various pre-training strategies.
However, existing methods are primarily designed for homophilic graphs and are largely limited to low-frequency information. This is due to their reliance on low-pass graph filters (e.g., GCN \cite{kipf2016semi}) and homophilic pre-training tasks such as link prediction \cite{sun2022gppt,liu2023graphprompt}, which are known to retain mainly low-frequency signals \cite{liu2022revisiting,yu2024non}. As a result, they can be viewed as special cases of our approach when restricted to low-frequency components.

Our spectral analysis reveals that in settings with sparse supervision, large spectral mismatch hinders effective alignment and parameter learning, which contributes to the failure of these homophily-based methods. 
While ProNoG \cite{yu2024non}, HeterGP \cite{yan2025hetergp}, and DAGPrompT \cite{chen2025dagprompt} have recently extended prompt tuning to heterophilic graphs via multi-view or distribution-aware strategies, they primarily operate in the spatial domain or rely on heuristic adaptations.
Crucially, they overlook the fundamental spectral mismatch between the frozen backbone and diverse downstream graphs. In contrast, our method explicitly treats prompt tuning as a spectral alignment process, enriching pre-trained knowledge and actively aligning the input spectrum to match pre-trained filters, ensuring robust performance even with limited labels.

\subsection{Heterophilic Graph Learning}
To address the heterophilic issue, many methods have been proposed, such as capturing high-frequency information \cite{bo2021beyond}, discovering potential neighbors \cite{jin2021node,pei2020geom}, and engaging in high-order message passing \cite{zhu2020beyond}. While they are trained under the supervised setting, there are also methods \cite{xiao2024simple,chen2024polygcl,pmlr-v235-wan24g} further aiming to design self-supervised learning for heterophilic graphs. These methods attempt to adopt more diverse neighbor selection approaches or learn high-frequency information \cite{chen2024polygcl,pmlr-v235-wan24g} through self-supervised tasks. However, these methods are tailored for full-model fine-tuning and do not provide a mechanism to transfer this knowledge to downstream tasks through lightweight prompts. Therefore, directly using the heterophilic priors can cause negative transfer, especially under limited supervision. Although graph prompt tuning holds the promise of enabling the transfer of pre-trained knowledge, the application of this paradigm to heterophilic graphs has been scarcely investigated. Our research commences in the spectral domain and conducts an in-depth understanding and exploration of graph prompt tuning specifically for heterophilic graphs. 

\section{Experiment Results}
In this section, we present the complete experimental results.

\subsection{Parameter Sensitivity}
\label{section:appendix_parameter}
\begin{figure}[t]
    \centering
    \includegraphics[scale=0.27]{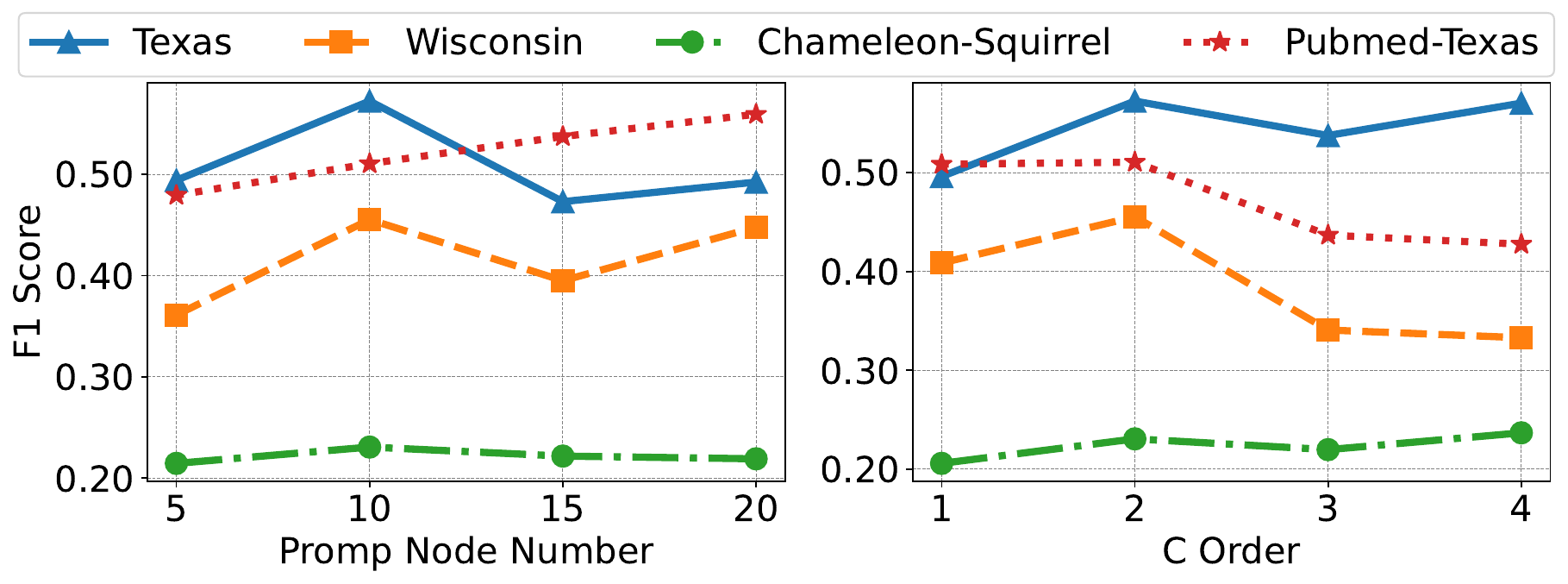}
    \vskip -1em
    \caption{Hyperparameter sensitivity evaluations regarding $C$ order and prompt node number.}
    \vskip -1em
    \label{fig:sensitivity_1}
\end{figure}
\begin{figure}[t]
    \centering
    \includegraphics[scale=0.27]{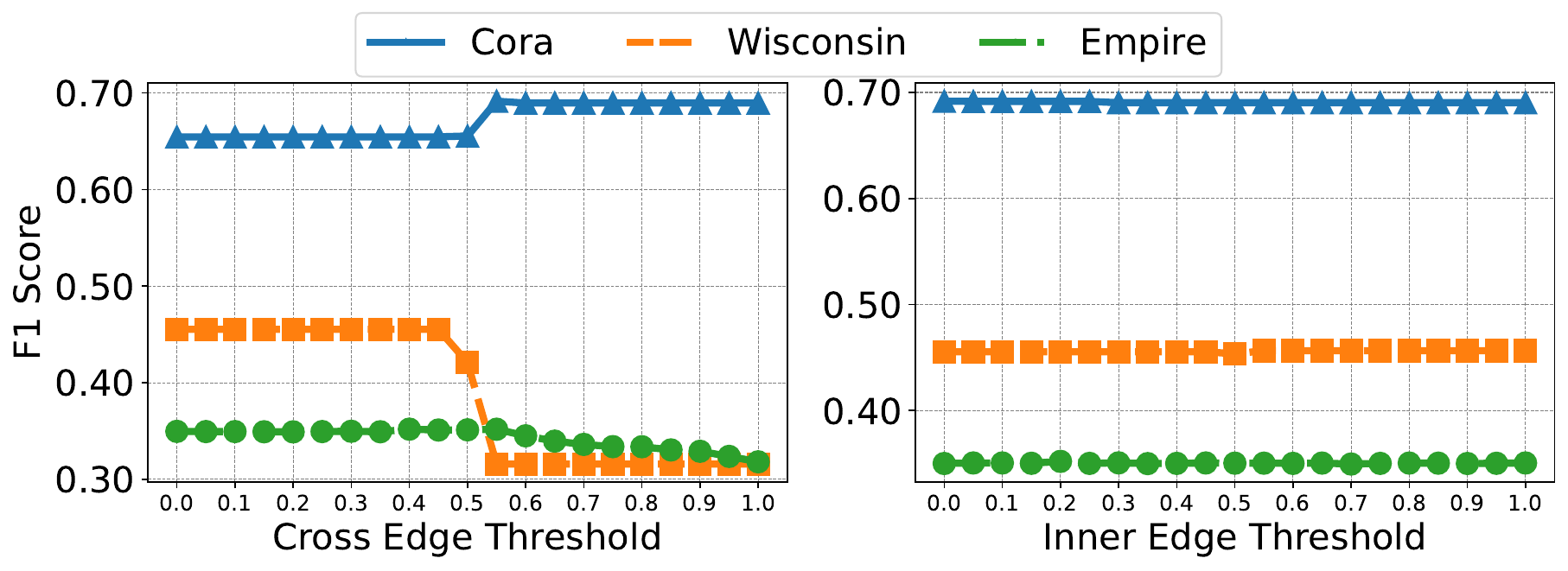}
    \vskip -1em
    \caption{Hyperparameter sensitivity evaluations regarding edge threshold.}
    \vskip -1em
    \label{fig:sensitivity_2}
\end{figure}

This section delves into the impact of key hyperparameters on the performance of our model. We adhere to the experimental setup described in Section \ref{section:performance_compar}.

Initially, we investigate two hyperparameters: the $C$ order and the number of nodes $N_p$ in each prompt graph. The results are presented in Figure \ref{fig:sensitivity_1}, from which we draw the following conclusions. (i) Increasing the number of prompt nodes initially enhances performance, enabling more adaptable spectral alignment. Yet, as the node count rises, the rate of improvement diminishes due to the scarcity of supervised data and elevated computational complexity. (ii) A comparable trend is observed for the $C$ order. Initially, more graph filters boost performance by integrating mode spectral knowledge, but the advantages wane as the number of filters increases. Given the limited amount of supervised data, an increase in model parameters complicates the model optimization process.

Subsequently, we examine how the thresholds $\tau_{inner}$ and $\tau_{cross}$ for establishing edges influence model performance. Specifically, we study both the thresholds for establishing cross edges and inner edges. During the experiments, we keep one type of threshold fixed while varying the other, and the results are illustrated in Figure \ref{fig:sensitivity_2}. The left sub-figure presents the results for cross edge thresholds, and the right one shows those for inner edge thresholds. The results uncover two significant findings: the performance is insensitive to inner edge thresholds, yet highly sensitive to cross edge thresholds. We now elaborate on these findings in greater detail.

(i) Regarding the threshold for cross edges, we set the inner edge threshold at 0.2 (in line with the previous work \cite{sun2023all}) and vary the cross edge threshold from 0 to 1. The model is evaluated on three datasets with varying degrees of homophily: Cora, Wisconsin, and Empire. The results reveal a distinct homophily-driven heuristic in the selection of the optimal threshold: the optimal cross edge threshold for homophilic graphs (e.g., Pubmed) is higher than that for heterophilic graphs (e.g., Wisconsin and Empire). This systematic variation stems from our dual objectives of maintaining structural integrity and ensuring the effectiveness of the prompts. In homophilic graphs, where neighboring nodes tend to be similar, we opt for a higher $\tau_{cross}$ value to ensure that connected nodes have more similar features and preserve the graph's homophilic nature. Conversely, in heterophilic graphs with more dissimilar neighboring nodes, we set a lower $\tau_{cross}$ to facilitate more diverse connections between prompt and original graph nodes.

(ii) For the threshold for inner edges, we set the cross edge threshold to 0.5 for Pubmed, and 0.4 for Wisconsin and Empire, and vary the inner edge threshold from 0 to 1. The results indicate that the performance is not sensitive to this threshold. We identify two primary reasons for this insensitivity. First, the number of prompt nodes is very small (10), making the internal topology governed by $\tau_{inner}$ structurally insignificant compared to the large downstream graphs. Second, the prompt node representations are learnable parameters optimized end-to-end. This allows the optimization process to adapt these representations to compensate for different inner-edge connectivities, further diminishing the hyperparameter's impact.

In summary, under our experimental setup, this reveals a clear distinction in sensitivity. The $\tau_{cross}$ threshold proves to be a critical parameter as it controls the primary mechanism for injecting learnable information into the main graph. Conversely, the $\tau_{inner}$ threshold is not critical because it only governs connections within a small, structurally secondary component whose influence can be absorbed by the adaptive learning process.

\subsection{Runtime Analysis}
\label{section:appendix_runtime}
\begin{table}[htbp]
    \centering
    \begin{tabular}{l|c|c|c}
        \toprule
        Runtime  & Cora & Texas & Ratings \\ \hline \hline
        GCN & 11.2971 & 10.4683 & 15.9558 \\ 
        GAT & 13.7155 & 12.6946 & 18.3710 \\ 
        BernNet & 25.7918 &25.4916	& 27.1818 \\
        ChebNetII & 43.4463	& 42.3719 & 43.9708 \\
        TFE-GNN & 10.0188 & 9.2063 & 13.1045 \\ \hline
        DGI & 15.1491 & 13.4188 & 15.1357 \\ 
        GraphCL & 15.0377 & 13.0939 & 15.0728 \\ 
        SimGRACE & 15.0573 & 13.0580 & 14.9969 \\ 
        PolyGCL & 31.197 & 30.1264 & 49.5764 \\ \hline
        GPPT & 886.2285 & 243.6319 & 90.3588 \\ 
        GPrompt & 80.925 & 26.59 & 1366.8791 \\ 
        GPF-plus & 80.6556 & 24.8727 & 1366.2642 \\ 
        All-in-One & 57.7804 & 36.909 & 180.2079 \\
        ProNoG & 360.1283 & 77.6541 & 6062.4575 \\ \hline
        HS-GPPT & 20.7673 & 20.1150 & 163.2120 \\ \bottomrule
    \end{tabular}
    \caption{Comparison of runtimes per epoch.}
    \label{tab:runtime}
\end{table}  

In this section, we carry out a runtime assessment of our model and the baseline models under the transductive learning settings specified in Section \ref{section:performance_compar}. We report the average time (measured in milliseconds, ms) consumed in each epoch, which includes both the training and validation processes. The results are presented in Table \ref{tab:runtime}, clearly demonstrating that our method has a relatively short runtime, thus guaranteeing high computational efficiency.

Among the baseline models, certain graph prompt methods (e.g., GPPT, GPrompt, and ProNoG) adopt similarity-based mechanisms for downstream predictions. This requires traversing all the nodes to acquire the corresponding class embeddings, which, as a result, significantly raises the time complexity. Moreover, in ProNoG, the readout from the downstream graph-specific condition net further exacerbates the time-complexity issue.

\subsection{Additional Accuracy Results}
\label{section:appendix_acc}
\begin{table*}
\centering
\scalebox{0.86}{
 \begin{tabular}{c|cc|cc|cc|cc|cc}
 \toprule
 Datasets    & \multicolumn{2}{c|}{Cora} & \multicolumn{2}{c|}{Pubmed} & \multicolumn{2}{c|}{Citeseer}  & \multicolumn{2}{c|}{Ratings} & \multicolumn{2}{c}{Empire} \\ \cline{1-11}
                        Metric      & Acc  & F1     & Acc    & F1 & Acc   & F1  & Acc  & F1 & Acc   & F1 \\ \hline \hline              
                        GCN     & \makecell{0.6532 \\ \footnotesize$\pm$\tiny{0.0319}} & \makecell{0.6506 \\ \footnotesize$\pm$\tiny{0.0185}} & \makecell{0.5546 \\ \footnotesize$\pm$\tiny{0.0367}} & \makecell{0.5405 \\ \footnotesize$\pm$\tiny{0.0358}} & \makecell{0.4442 \\ \footnotesize$\pm$\tiny{0.0303}} & \makecell{0.4187 \\ \footnotesize$\pm$\tiny{0.0220}}  & \makecell{0.2384 \\ \footnotesize$\pm$\tiny{0.0150}} & \makecell{0.1911 \\ \footnotesize$\pm$\tiny{0.0634}} & \makecell{0.1613 \\ \footnotesize$\pm$\tiny{0.0145}} & \makecell{0.1491 \\ \footnotesize$\pm$\tiny{0.0707}} \\
                        GAT & \makecell{0.6018 \\ \footnotesize$\pm$\tiny{0.0238}} & \makecell{0.6091 \\ \footnotesize$\pm$\tiny{0.0201}} & \makecell{0.5472 \\ \footnotesize$\pm$\tiny{0.0641}} & \makecell{0.5381 \\ \footnotesize$\pm$\tiny{0.0595}} & \makecell{0.4446 \\ \footnotesize$\pm$\tiny{0.0350}} & \makecell{0.4162 \\ \footnotesize$\pm$\tiny{0.0332}} & \makecell{0.2002 \\ \footnotesize$\pm$\tiny{0.0248}} & \makecell{0.1827 \\ \footnotesize$\pm$\tiny{0.0371}} & \makecell{0.1419 \\ \footnotesize$\pm$\tiny{0.0221}} & \makecell{0.1367 \\ \footnotesize$\pm$\tiny{0.0208}}   \\
                        BernNet & \makecell{0.6494 \\ \footnotesize$\pm$\tiny{0.0280}} & \makecell{0.6356 \\ \footnotesize$\pm$\tiny{0.0296}} & \makecell{0.6584 \\ \footnotesize$\pm$\tiny{0.0216}} & \makecell{0.6493 \\ \footnotesize$\pm$\tiny{0.0215}} & \makecell{0.4888 \\ \footnotesize$\pm$\tiny{0.0197}} & \makecell{0.4712 \\ \footnotesize$\pm$\tiny{0.0192}} & \makecell{0.2581 \\ \footnotesize$\pm$\tiny{0.0074}} & \textbf{\makecell{0.2030 \\ \footnotesize$\pm$\tiny{0.0063}}} & \makecell{0.1215 \\ \footnotesize$\pm$\tiny{0.0280}} & \makecell{0.1071 \\ \footnotesize$\pm$\tiny{0.0253}}  \\
                        ChebNetII & \makecell{0.6798 \\ \footnotesize$\pm$\tiny{0.0170}} & \makecell{0.6752 \\ \footnotesize$\pm$\tiny{0.0202}} & \makecell{0.6616 \\ \footnotesize$\pm$\tiny{0.0431}} & \makecell{0.6554 \\ \footnotesize$\pm$\tiny{0.0410}} & \makecell{0.4526 \\ \footnotesize$\pm$\tiny{0.0353}} & \makecell{0.4304 \\ \footnotesize$\pm$\tiny{0.0328}} & \makecell{0.2382 \\ \footnotesize$\pm$\tiny{0.0148}} & \makecell{0.1970 \\ \footnotesize$\pm$\tiny{0.0047}} & \makecell{0.0661 \\ \footnotesize$\pm$\tiny{0.0088}} & \makecell{0.0551 \\ \footnotesize$\pm$\tiny{0.0054}}   \\
                        TFE-GNN  & \makecell{0.3376 \\ \footnotesize$\pm$\tiny{0.0198}} & \makecell{0.3286 \\ \footnotesize$\pm$\tiny{0.0177}} & \makecell{0.4694 \\ \footnotesize$\pm$\tiny{0.0170}} & \makecell{0.4590 \\ \footnotesize$\pm$\tiny{0.0155}} & \makecell{0.2932 \\ \footnotesize$\pm$\tiny{0.0237}} & \makecell{0.2874 \\ \footnotesize$\pm$\tiny{0.0203}} & \makecell{0.2301 \\ \footnotesize$\pm$\tiny{0.0099}} & \makecell{0.1939 \\ \footnotesize$\pm$\tiny{0.0014}} & \makecell{\underline{0.3502} \\ \footnotesize$\pm$\tiny{0.0151}} & \makecell{\underline{0.3155} \\ \footnotesize$\pm$\tiny{0.0182}} \\
                        \cline{1-11}
                        DGI & \makecell{0.3658 \\ \footnotesize$\pm$\tiny{0.1487}} & \makecell{0.3187 \\ \footnotesize$\pm$\tiny{0.1251}} & \makecell{0.4870 \\ \footnotesize$\pm$\tiny{0.0834}} & \makecell{0.3743 \\ \footnotesize$\pm$\tiny{0.1073}} & \makecell{0.2922 \\ \footnotesize$\pm$\tiny{0.0347}} & \makecell{0.2235 \\ \footnotesize$\pm$\tiny{0.0405}} & \makecell{0.1818 \\ \footnotesize$\pm$\tiny{0.0418}} & \makecell{0.1485 \\ \footnotesize$\pm$\tiny{0.0237}} & \makecell{0.1209 \\ \footnotesize$\pm$\tiny{0.0359}} & \makecell{0.0710 \\ \footnotesize$\pm$\tiny{0.0188}} \\
                        GraphCL & \makecell{0.5958 \\ \footnotesize$\pm$\tiny{0.0395}} & \makecell{0.5603 \\ \footnotesize$\pm$\tiny{0.0486}} & \makecell{0.5694 \\ \footnotesize$\pm$\tiny{0.0327}} & \makecell{0.5576 \\ \footnotesize$\pm$\tiny{0.0396}} & \makecell{0.4154 \\ \footnotesize$\pm$\tiny{0.0263}} & \makecell{0.3741 \\ \footnotesize$\pm$\tiny{0.0356}} & \makecell{0.2326 \\ \footnotesize$\pm$\tiny{0.0358}} & \makecell{0.1853 \\ \footnotesize$\pm$\tiny{0.0158}} & \makecell{0.1184 \\ \footnotesize$\pm$\tiny{0.0245}} & \makecell{0.0961 \\ \footnotesize$\pm$\tiny{0.0150}}  \\
                        SimGRACE & \makecell{0.4400 \\ \footnotesize$\pm$\tiny{0.0321}} & \makecell{0.4283 \\ \footnotesize$\pm$\tiny{0.0341}} & \makecell{0.4416 \\ \footnotesize$\pm$\tiny{0.0424}} & \makecell{0.4316 \\ \footnotesize$\pm$\tiny{0.0482}} & \makecell{0.3664 \\ \footnotesize$\pm$\tiny{0.0392}} & \makecell{0.3412 \\ \footnotesize$\pm$\tiny{0.0405}} & \makecell{0.2064 \\ \footnotesize$\pm$\tiny{0.0239}} & \makecell{0.1761 \\ \footnotesize$\pm$\tiny{0.0153}} & \makecell{0.0878 \\ \footnotesize$\pm$\tiny{0.0097}} & \makecell{0.0635 \\ \footnotesize$\pm$\tiny{0.0065}} \\
                        PolyGCL  & \makecell{0.6948 \\ \footnotesize$\pm$\tiny{0.0284}} & \makecell{\underline{0.6655} \\ \footnotesize$\pm$\tiny{0.0221}} & \makecell{\underline{0.6792} \\ \footnotesize$\pm$\tiny{0.0364}} & \makecell{\underline{0.6782} \\ \footnotesize$\pm$\tiny{0.0357}} & \textbf{\makecell{0.5266 \\ \footnotesize$\pm$\tiny{0.0286}}} & \makecell{\underline{0.5010} \\ \footnotesize$\pm$\tiny{0.0224}} & \makecell{0.2048 \\ \footnotesize$\pm$\tiny{0.0155}} & \makecell{0.1874 \\ \footnotesize$\pm$\tiny{0.0081}} & \makecell{0.0634 \\ \footnotesize$\pm$\tiny{0.0060}} & \makecell{0.0528 \\ \footnotesize$\pm$\tiny{0.0029}} \\
                        \cline{1-11}
                        GPPT & \makecell{\underline{0.7026} \\ \footnotesize$\pm$\tiny{0.0172}} & \makecell{0.6611 \\ \footnotesize$\pm$\tiny{0.0313}} & \makecell{0.6540 \\ \footnotesize$\pm$\tiny{0.0452}} & \makecell{0.6459 \\ \footnotesize$\pm$\tiny{0.0447}} & \makecell{0.3854 \\ \footnotesize$\pm$\tiny{0.0617}} & \makecell{0.3731 \\ \footnotesize$\pm$\tiny{0.0608}} & \makecell{0.2345 \\ \footnotesize$\pm$\tiny{0.0156}} & \makecell{0.1944 \\ \footnotesize$\pm$\tiny{0.0128}} & \makecell{0.0750 \\ \footnotesize$\pm$\tiny{0.0037}} & \makecell{0.0658 \\ \footnotesize$\pm$\tiny{0.0036}} \\
                        GPrompt  & \makecell{0.6578 \\ \footnotesize$\pm$\tiny{0.0323}} & \makecell{0.6366 \\ \footnotesize$\pm$\tiny{0.0252}} & \makecell{0.6230 \\ \footnotesize$\pm$\tiny{0.0428}} &  \makecell{0.6113 \\ \footnotesize$\pm$\tiny{0.0428}} & \makecell{0.4568 \\ \footnotesize$\pm$\tiny{0.0269}} & \makecell{0.4316 \\ \footnotesize$\pm$\tiny{0.0315}} & \makecell{0.2269 \\ \footnotesize$\pm$\tiny{0.0252}} & \makecell{0.1946 \\ \footnotesize$\pm$\tiny{0.0125}} & \makecell{0.0652 \\ \footnotesize$\pm$\tiny{0.0071}} & \makecell{0.0539 \\ \footnotesize$\pm$\tiny{0.0040}} \\
                        GPF-plus  & \makecell{0.6250 \\ \footnotesize$\pm$\tiny{0.0279}} & \makecell{0.6160 \\ \footnotesize$\pm$\tiny{0.0251}} & \makecell{0.6706 \\ \footnotesize$\pm$\tiny{0.0242}} &  \makecell{0.6534 \\ \footnotesize$\pm$\tiny{0.0250}} & \makecell{0.4564 \\ \footnotesize$\pm$\tiny{0.0308}} & \makecell{0.4441 \\ \footnotesize$\pm$\tiny{0.0322}} & \makecell{0.2092 \\ \footnotesize$\pm$\tiny{0.0273}} & \makecell{0.1826 \\ \footnotesize$\pm$\tiny{0.0177}} & \makecell{0.0655 \\ \footnotesize$\pm$\tiny{0.0080}} & \makecell{0.0476 \\ \footnotesize$\pm$\tiny{0.0055}} \\
                        All-in-One  & \makecell{0.5442 \\ \footnotesize$\pm$\tiny{0.0731}} & \makecell{0.4308 \\ \footnotesize$\pm$\tiny{0.0410}} & \makecell{0.6724 \\ \footnotesize$\pm$\tiny{0.0335}} &  \makecell{0.6600 \\ \footnotesize$\pm$\tiny{0.0330}} & \makecell{0.4616 \\ \footnotesize$\pm$\tiny{0.0095}} & \makecell{0.3638 \\ \footnotesize$\pm$\tiny{0.0158}} & \textbf{\makecell{0.2679 \\ \footnotesize$\pm$\tiny{0.0371}}} & \makecell{0.1819 \\ \footnotesize$\pm$\tiny{0.0263}} & \makecell{0.0566 \\ \footnotesize$\pm$\tiny{0.0037}} & \makecell{0.0472 \\ \footnotesize$\pm$\tiny{0.0028}}   \\
                        ProNoG  & \makecell{0.5838 \\ \footnotesize$\pm$\tiny{0.0121}} & \makecell{0.5564 \\ \footnotesize$\pm$\tiny{0.0203}} & \makecell{0.5348 \\ \footnotesize$\pm$\tiny{0.0748}} & \makecell{0.5242 \\ \footnotesize$\pm$\tiny{0.0855}} & \makecell{0.2596 \\ \footnotesize$\pm$\tiny{0.0160}} & \makecell{0.2466 \\ \footnotesize$\pm$\tiny{0.0172}} & \makecell{\underline{0.2513} \\ \footnotesize$\pm$\tiny{0.0426}} & \makecell{0.1963 \\ \footnotesize$\pm$\tiny{0.0102}} & \makecell{0.1025 \\ \footnotesize$\pm$\tiny{0.0142}} & \makecell{0.0784 \\ \footnotesize$\pm$\tiny{0.0090}} \\
                        \cline{1-11}
                        HS-GPPT  & \textbf{\makecell{0.7156 \\ \footnotesize$\pm$\tiny{0.0132}}} & \textbf{\makecell{0.6915 \\ \footnotesize$\pm$\tiny{0.0166}}} & \textbf{\makecell{0.6912 \\ \footnotesize$\pm$\tiny{0.0215}}} & \textbf{\makecell{0.6910 \\ \footnotesize$\pm$\tiny{0.0199}}} & \makecell{\underline{0.5188} \\ \footnotesize$\pm$\tiny{0.0351}} & \textbf{\makecell{0.5043 \\ \footnotesize$\pm$\tiny{0.0223}}} & \makecell{0.2407 \\ \footnotesize$\pm$\tiny{0.0239}} & \makecell{\underline{0.1972} \\ \footnotesize$\pm$\tiny{0.0123}} & \textbf{\makecell{0.3836 \\ \footnotesize$\pm$\tiny{0.0322}}} & \textbf{\makecell{0.3520 \\ \footnotesize$\pm$\tiny{0.0194}}} 
                         \\ \bottomrule
 \end{tabular}}
\caption{Accuracy and F1 score on homophilic graphs and large-scale heterophilic graphs under transductive learning.}
\label{tab:appendix_homo_trans}
\end{table*}

\begin{table*}
\centering
\scalebox{0.86}{
 \begin{tabular}{c|cc|cc|cc|cc|cc}
 \toprule
 Datasets    & \multicolumn{2}{c|}{Cornell} & \multicolumn{2}{c|}{Texas} & \multicolumn{2}{c|}{Wisconsin} & \multicolumn{2}{c|}{Chameleon}  & \multicolumn{2}{c}{Squirrel}  \\ \cline{1-11}
                        Metric      & Acc  & F1     & Acc    & F1 & Acc   & F1 & Acc &F1  & Acc &F1 \\ \hline \hline              
                        GCN     & \makecell{0.2535 \\ \footnotesize$\pm$\tiny{0.0611}} & \makecell{0.1835 \\ \footnotesize$\pm$\tiny{0.0302}} & \makecell{0.4028 \\ \footnotesize$\pm$\tiny{0.1002}} & \makecell{0.2506 \\ \footnotesize$\pm$\tiny{0.0571}} & \makecell{0.3782 \\ \footnotesize$\pm$\tiny{0.0853}} & \makecell{0.2496 \\ \footnotesize$\pm$\tiny{0.0411}} & \makecell{0.3117 \\ \footnotesize$\pm$\tiny{0.0532}} & \makecell{0.2998 \\ \footnotesize$\pm$\tiny{0.0510}} & \makecell{\underline{0.2541} \\ \footnotesize$\pm$\tiny{0.0123}} & \makecell{0.2441 \\ \footnotesize$\pm$\tiny{0.0141}} \\
                        GAT & \makecell{0.2620 \\ \footnotesize$\pm$\tiny{0.1373}} & \makecell{0.1531 \\ \footnotesize$\pm$\tiny{0.0747}} & \makecell{0.2986 \\ \footnotesize$\pm$\tiny{0.0242}} & \makecell{0.2144 \\ \footnotesize$\pm$\tiny{0.0471}} & \makecell{0.2891 \\ \footnotesize$\pm$\tiny{0.0961}} & \makecell{0.1836 \\ \footnotesize$\pm$\tiny{0.0463}}  & \makecell{0.2989 \\ \footnotesize$\pm$\tiny{0.0489}}  & \makecell{0.2873 \\ \footnotesize$\pm$\tiny{0.0462}} & \makecell{0.2307 \\ \footnotesize$\pm$\tiny{0.0067}} & \makecell{0.2281 \\ \footnotesize$\pm$\tiny{0.0016}}   \\
                        BernNet & \makecell{0.3042 \\ \footnotesize$\pm$\tiny{0.0727}} & \makecell{0.2287 \\ \footnotesize$\pm$\tiny{0.0393}} & \makecell{0.5915 \\ \footnotesize$\pm$\tiny{0.0309}} & \makecell{0.3743 \\ \footnotesize$\pm$\tiny{0.0476}} & \makecell{0.2950 \\ \footnotesize$\pm$\tiny{0.0536}} & \makecell{0.2215 \\ \footnotesize$\pm$\tiny{0.0343}} & \makecell{0.2785 \\ \footnotesize$\pm$\tiny{0.0134}} & \makecell{0.2745 \\ \footnotesize$\pm$\tiny{0.0139}} & \makecell{0.2161 \\ \footnotesize$\pm$\tiny{0.0112}} & \makecell{0.2150 \\ \footnotesize$\pm$\tiny{0.0108}}  \\
                        ChebNetII & \makecell{0.3296 \\ \footnotesize$\pm$\tiny{0.0433}} & \makecell{0.2080 \\ \footnotesize$\pm$\tiny{0.0288}} & \makecell{0.5606 \\ \footnotesize$\pm$\tiny{0.0559}} & \makecell{0.3558 \\ \footnotesize$\pm$\tiny{0.0487}} & \makecell{0.3584 \\ \footnotesize$\pm$\tiny{0.0462}} & \makecell{0.2533 \\ \footnotesize$\pm$\tiny{0.0525}} & \makecell{0.2718 \\ \footnotesize$\pm$\tiny{0.0247}} & \makecell{0.2684 \\ \footnotesize$\pm$\tiny{0.0265}} & \makecell{0.2144 \\ \footnotesize$\pm$\tiny{0.0123}} & \makecell{0.2133 \\ \footnotesize$\pm$\tiny{0.0123}}   \\
                        TFE-GNN & \makecell{\underline{0.4535} \\ \footnotesize$\pm$\tiny{0.1001}} & \textbf{\makecell{0.3950 \\ \footnotesize$\pm$\tiny{0.1131}}} & \makecell{0.4563 \\ \footnotesize$\pm$\tiny{0.1230}} & \makecell{0.3730 \\ \footnotesize$\pm$\tiny{0.1375}} & \makecell{\underline{0.4733} \\ \footnotesize$\pm$\tiny{0.0973}} & \makecell{\underline{0.3828} \\ \footnotesize$\pm$\tiny{0.0822}} & \makecell{0.2907 \\ \footnotesize$\pm$\tiny{0.0091}} & \makecell{0.2872 \\ \footnotesize$\pm$\tiny{0.0115}} & \makecell{0.2256 \\ \footnotesize$\pm$\tiny{0.0056}} & \makecell{0.2242 \\ \footnotesize$\pm$\tiny{0.0050}} \\
                        \cline{1-11}
                        DGI & \makecell{0.2648 \\ \footnotesize$\pm$\tiny{0.0552}} & \makecell{0.1650 \\ \footnotesize$\pm$\tiny{0.0311}} & \makecell{0.4789 \\ \footnotesize$\pm$\tiny{0.1321}} & \makecell{0.3031 \\ \footnotesize$\pm$\tiny{0.0804}} & \makecell{0.3545 \\ \footnotesize$\pm$\tiny{0.0967}} & \makecell{0.2293 \\ \footnotesize$\pm$\tiny{0.0422}}  & \makecell{0.2907 \\ \footnotesize$\pm$\tiny{0.0388}} & \makecell{0.2692 \\ \footnotesize$\pm$\tiny{0.0336}} & \makecell{0.2274 \\ \footnotesize$\pm$\tiny{0.0301}} & \makecell{0.1989 \\ \footnotesize$\pm$\tiny{0.0195}} \\
                        GraphCL & \makecell{0.2648 \\ \footnotesize$\pm$\tiny{0.1258}} & \makecell{0.1826 \\ \footnotesize$\pm$\tiny{0.0579}} & \makecell{0.4676 \\ \footnotesize$\pm$\tiny{0.0833}} & \makecell{0.2673 \\ \footnotesize$\pm$\tiny{0.0522}} & \makecell{0.3386 \\ \footnotesize$\pm$\tiny{0.0731}} & \makecell{0.2414 \\ \footnotesize$\pm$\tiny{0.0313}} & \makecell{0.2921 \\ \footnotesize$\pm$\tiny{0.0475}} & \makecell{0.2710 \\ \footnotesize$\pm$\tiny{0.0432}} & \makecell{0.2215 \\ \footnotesize$\pm$\tiny{0.0066}} & \makecell{0.1890 \\ \footnotesize$\pm$\tiny{0.0129}} \\
                        SimGRACE & \makecell{0.3296 \\ \footnotesize$\pm$\tiny{0.0738}} & \makecell{0.1904 \\ \footnotesize$\pm$\tiny{0.0237}} & \makecell{0.4732 \\ \footnotesize$\pm$\tiny{0.0423}} & \makecell{0.2833 \\ \footnotesize$\pm$\tiny{0.0187}} & \makecell{0.3129 \\ \footnotesize$\pm$\tiny{0.0926}} & \makecell{0.2116 \\ \footnotesize$\pm$\tiny{0.0625}} & \makecell{0.3022 \\ \footnotesize$\pm$\tiny{0.0275}} & \makecell{0.2706 \\ \footnotesize$\pm$\tiny{0.0228}} & \makecell{0.2261 \\ \footnotesize$\pm$\tiny{0.0164}} & \makecell{0.2071 \\ \footnotesize$\pm$\tiny{0.0165}}  \\
                        PolyGCL & \makecell{0.3070 \\ \footnotesize$\pm$\tiny{0.0559}} & \makecell{0.2268 \\ \footnotesize$\pm$\tiny{0.0469}} & \makecell{\underline{0.6338} \\ \footnotesize$\pm$\tiny{0.0570}} & \makecell{\underline{0.4913} \\ \footnotesize$\pm$\tiny{0.0568}} & \makecell{0.3644 \\ \footnotesize$\pm$\tiny{0.0820}} & \makecell{0.2254 \\ \footnotesize$\pm$\tiny{0.0273}} & \textbf{\makecell{0.3415 \\ \footnotesize$\pm$\tiny{0.0379}}} & \makecell{\underline{0.3308} \\ \footnotesize$\pm$\tiny{0.0351}} & \makecell{0.2516 \\ \footnotesize$\pm$\tiny{0.0170}} & \makecell{\underline{0.2450} \\ \footnotesize$\pm$\tiny{0.0129}} \\
                        \cline{1-11}
                        GPPT & \makecell{0.3352 \\ \footnotesize$\pm$\tiny{0.0491}} & \makecell{0.1798 \\ \footnotesize$\pm$\tiny{0.0447}} & \makecell{0.5634 \\ \footnotesize$\pm$\tiny{0.0527}} & \makecell{0.3247 \\ \footnotesize$\pm$\tiny{0.0407}} & \makecell{0.3426 \\ \footnotesize$\pm$\tiny{0.0971}} & \makecell{0.2619 \\ \footnotesize$\pm$\tiny{0.0726}} & \makecell{0.3092 \\ \footnotesize$\pm$\tiny{0.0191}} & \makecell{0.2976 \\ \footnotesize$\pm$\tiny{0.0221}} & \makecell{0.2163 \\ \footnotesize$\pm$\tiny{0.0142}} & \makecell{0.2082 \\ \footnotesize$\pm$\tiny{0.0187}} \\
                        GPrompt  & \makecell{0.3437 \\ \footnotesize$\pm$\tiny{0.1262}} & \makecell{0.1918 \\ \footnotesize$\pm$\tiny{0.0365}} & \makecell{0.3634 \\ \footnotesize$\pm$\tiny{0.2001}} &  \makecell{0.1932 \\ \footnotesize$\pm$\tiny{0.0846}} & \makecell{0.3505 \\ \footnotesize$\pm$\tiny{0.0860}} & \makecell{0.1989 \\ \footnotesize$\pm$\tiny{0.0425}} & \makecell{0.2539 \\ \footnotesize$\pm$\tiny{0.0159}} & \makecell{0.2326 \\ \footnotesize$\pm$\tiny{0.0144}} & \makecell{0.2088 \\ \footnotesize$\pm$\tiny{0.0197}} & \makecell{0.2019 \\ \footnotesize$\pm$\tiny{0.0176}} \\
                        GPF-plus  & \makecell{0.3549 \\ \footnotesize$\pm$\tiny{0.1629}} & \makecell{0.1956 \\ \footnotesize$\pm$\tiny{0.0716}} & \makecell{0.4761 \\ \footnotesize$\pm$\tiny{0.0449}} &  \makecell{0.2515 \\ \footnotesize$\pm$\tiny{0.0353}} & \makecell{0.2891 \\ \footnotesize$\pm$\tiny{0.1204}} & \makecell{0.1928 \\ \footnotesize$\pm$\tiny{0.0688}} & \makecell{0.2467 \\ \footnotesize$\pm$\tiny{0.0246}} & \makecell{0.2317 \\ \footnotesize$\pm$\tiny{0.0245}} & \makecell{0.2076 \\ \footnotesize$\pm$\tiny{0.0121}} & \makecell{0.1844 \\ \footnotesize$\pm$\tiny{0.0187}} \\
                        All-in-One  & \makecell{0.3662 \\ \footnotesize$\pm$\tiny{0.1718}} & \makecell{0.1479 \\ \footnotesize$\pm$\tiny{0.0554}} & \makecell{0.4028 \\ \footnotesize$\pm$\tiny{0.0250}} &  \makecell{0.1825 \\ \footnotesize$\pm$\tiny{0.0650}} & \makecell{0.3069 \\ \footnotesize$\pm$\tiny{0.1693}} & \makecell{0.1381 \\ \footnotesize$\pm$\tiny{0.0731}} & \makecell{0.2417 \\ \footnotesize$\pm$\tiny{0.0180}} & \makecell{0.2254 \\ \footnotesize$\pm$\tiny{0.0164}} & \makecell{0.2249 \\ \footnotesize$\pm$\tiny{0.0530}} & \makecell{0.1751 \\ \footnotesize$\pm$\tiny{0.0324}} \\
                        ProNoG & \makecell{0.2563 \\ \footnotesize$\pm$\tiny{0.0449}} & \makecell{0.1987 \\ \footnotesize$\pm$\tiny{0.0289}} & \makecell{0.5070 \\ \footnotesize$\pm$\tiny{0.0807}} & \makecell{0.2627 \\ \footnotesize$\pm$\tiny{0.0420}} & \makecell{0.3366 \\ \footnotesize$\pm$\tiny{0.0349}} & \makecell{0.2218 \\ \footnotesize$\pm$\tiny{0.0371}} & \makecell{0.2803 \\ \footnotesize$\pm$\tiny{0.0296}} & \makecell{0.2565 \\ \footnotesize$\pm$\tiny{0.0294}} & \makecell{0.2118 \\ \footnotesize$\pm$\tiny{0.0072}} & \makecell{0.2030 \\ \footnotesize$\pm$\tiny{0.0049}} \\
                        \cline{1-11}
                        HS-GPPT & \textbf{\makecell{0.5239 \\ \footnotesize$\pm$\tiny{0.0675}}} & \makecell{\underline{0.3875} \\ \footnotesize$\pm$\tiny{0.0777}} & \textbf{\makecell{0.6423 \\ \footnotesize$\pm$\tiny{0.1041}}} & \textbf{\makecell{0.5724 \\ \footnotesize$\pm$\tiny{0.0446}}} & \textbf{\makecell{0.5683 \\ \footnotesize$\pm$\tiny{0.0640}}} & \textbf{\makecell{0.4554 \\ \footnotesize$\pm$\tiny{0.0484}}} & \makecell{\underline{0.3302} \\ \footnotesize$\pm$\tiny{0.0357}} & \textbf{\makecell{0.3324 \\ \footnotesize$\pm$\tiny{0.0360}}} & \textbf{\makecell{0.2589 \\ \footnotesize$\pm$\tiny{0.0144}}} & \textbf{\makecell{0.2536 \\ \footnotesize$\pm$\tiny{0.0108}}}  \\ \bottomrule
 \end{tabular}}
 \caption{Accuracy and F1 score on heterophilic graphs under transductive learning.}
\label{tab:appendix_hete_trans}
\end{table*}

\begin{table*}
\centering
\scalebox{0.73}{
 \begin{tabular}{c|cc|cc|cc|cc|cc|cc}
 \toprule
 Source     & \multicolumn{2}{c|}{Texas} & \multicolumn{2}{c|}{Wisconsin}  & \multicolumn{2}{c|}{Chameleon} & \multicolumn{2}{c|}{Pubmed}  & \multicolumn{2}{c|}{Squirrel}  & \multicolumn{2}{c}{Ratings} \\ 
  Target    & \multicolumn{2}{c|}{Wisconsin} & \multicolumn{2}{c|}{Texas}  & \multicolumn{2}{c|}{Squirrel} & \multicolumn{2}{c|}{Texas}  & \multicolumn{2}{c|}{Cornell}  & \multicolumn{2}{c}{Empire} \\ \cline{1-13}
                        Metric      & Acc  & F1     & Acc    & F1 & Acc   & F1 & Acc &F1  & Acc &F1  & Acc &F1 \\ \hline \hline           
                        DGI & \makecell{0.3485 \\ \footnotesize$\pm$\tiny{0.1185}} & \makecell{0.2123 \\ \footnotesize$\pm$\tiny{0.0627}} & \makecell{0.3408 \\ \footnotesize$\pm$\tiny{0.1059}} & \makecell{0.1976 \\ \footnotesize$\pm$\tiny{0.0721}}  & \makecell{0.2240 \\ \footnotesize$\pm$\tiny{0.0142}} & \makecell{0.2048 \\ \footnotesize$\pm$\tiny{0.0093}}  & \makecell{0.4423 \\ \footnotesize$\pm$\tiny{0.1526}} & \makecell{0.2177 \\ \footnotesize$\pm$\tiny{0.0762}}  & \makecell{0.2423 \\ \footnotesize$\pm$\tiny{0.0402}} & \makecell{0.1450 \\ \footnotesize$\pm$\tiny{0.0258}} & \makecell{\underline{0.1008} \\ \footnotesize$\pm$\tiny{0.0204}} & \makecell{0.0615 \\ \footnotesize$\pm$\tiny{0.0075}} \\
                        GraphCL & \makecell{0.1941 \\ \footnotesize$\pm$\tiny{0.0495}} & \makecell{0.1433 \\ \footnotesize$\pm$\tiny{0.0313}} & \makecell{0.5127 \\ \footnotesize$\pm$\tiny{0.1274}} & \makecell{0.2847 \\ \footnotesize$\pm$\tiny{0.0680}}  & \makecell{0.2245 \\ \footnotesize$\pm$\tiny{0.0198}} & \makecell{0.2098 \\ \footnotesize$\pm$\tiny{0.0164}} & \makecell{0.4366 \\ \footnotesize$\pm$\tiny{0.0782}} & \makecell{0.2684 \\ \footnotesize$\pm$\tiny{0.0266}}  & \makecell{0.2282 \\ \footnotesize$\pm$\tiny{0.0620}} & \makecell{0.1686 \\ \footnotesize$\pm$\tiny{0.0438}} & \makecell{0.0990 \\ \footnotesize$\pm$\tiny{0.0199}} & \makecell{0.0681 \\ \footnotesize$\pm$\tiny{0.0090}}  \\
                        SimGRACE & \makecell{0.3228 \\ \footnotesize$\pm$\tiny{0.0774}} & \makecell{0.1957 \\ \footnotesize$\pm$\tiny{0.0294}} & \makecell{0.3465 \\ \footnotesize$\pm$\tiny{0.1293}} & \makecell{0.2309 \\ \footnotesize$\pm$\tiny{0.0499}} & \makecell{\underline{0.2276} \\ \footnotesize$\pm$\tiny{0.0140}} & \makecell{0.2111 \\ \footnotesize$\pm$\tiny{0.0141}} & \makecell{0.3408 \\ \footnotesize$\pm$\tiny{0.1074}} & \makecell{0.1834 \\ \footnotesize$\pm$\tiny{0.0438}}  & \makecell{0.1972 \\ \footnotesize$\pm$\tiny{0.0891}} & \makecell{0.1369 \\ \footnotesize$\pm$\tiny{0.0720}}  & \makecell{0.0958 \\ \footnotesize$\pm$\tiny{0.0186}} & \makecell{0.0753 \\ \footnotesize$\pm$\tiny{0.0098}} \\
                        PolyGCL & \makecell{\underline{0.3960} \\ \footnotesize$\pm$\tiny{0.0689}} & \makecell{\underline{0.2473} \\ \footnotesize$\pm$\tiny{0.0460}} & \makecell{\underline{0.5549} \\ \footnotesize$\pm$\tiny{0.0423}} & \makecell{\underline{0.3901} \\ \footnotesize$\pm$\tiny{0.0585}}  & \makecell{0.2218 \\ \footnotesize$\pm$\tiny{0.0213}} & \makecell{\underline{0.2130} \\ \footnotesize$\pm$\tiny{0.0217}} & \makecell{0.4366 \\ \footnotesize$\pm$\tiny{0.1061}} & \makecell{0.2934 \\ \footnotesize$\pm$\tiny{0.0774}}  & \makecell{0.3324 \\ \footnotesize$\pm$\tiny{0.1115}} & \makecell{0.2112 \\ \footnotesize$\pm$\tiny{0.0410}} & \makecell{0.0593 \\ \footnotesize$\pm$\tiny{0.0043}} & \makecell{0.0487 \\ \footnotesize$\pm$\tiny{0.0020}}  \\
                        \cline{1-13}
                        GPPT & \makecell{0.3683 \\ \footnotesize$\pm$\tiny{0.0637}} & \makecell{0.2295 \\ \footnotesize$\pm$\tiny{0.0564}} & \makecell{0.5437 \\ \footnotesize$\pm$\tiny{0.0211}} & \makecell{0.3774 \\ \footnotesize$\pm$\tiny{0.0678}}  & \makecell{0.2155 \\ \footnotesize$\pm$\tiny{0.0167}} & \makecell{0.1987 \\ \footnotesize$\pm$\tiny{0.0189}} & \makecell{\underline{0.5437} \\ \footnotesize$\pm$\tiny{0.0211}} & \makecell{\underline{0.3774} \\ \footnotesize$\pm$\tiny{0.0678}}  & \makecell{0.2282 \\ \footnotesize$\pm$\tiny{0.0522}} & \makecell{0.1906 \\ \footnotesize$\pm$\tiny{0.0435}} & \makecell{0.0990 \\ \footnotesize$\pm$\tiny{0.0155}} & \makecell{\underline{0.0860} \\ \footnotesize$\pm$\tiny{0.0049}}   \\
                        GPrompt & \makecell{0.2911 \\ \footnotesize$\pm$\tiny{0.1204}} & \makecell{0.1960 \\ \footnotesize$\pm$\tiny{0.0614}} & \makecell{0.3662 \\ \footnotesize$\pm$\tiny{0.1336}} & \makecell{0.2078 \\ \footnotesize$\pm$\tiny{0.0608}} & \makecell{0.2042 \\ \footnotesize$\pm$\tiny{0.0125}} & \makecell{0.1982 \\ \footnotesize$\pm$\tiny{0.0091}} & \makecell{0.3972 \\ \footnotesize$\pm$\tiny{0.1467}} & \makecell{0.2131 \\ \footnotesize$\pm$\tiny{0.0386}}  & \makecell{0.3408 \\ \footnotesize$\pm$\tiny{0.1139}} & \makecell{0.1825 \\ \footnotesize$\pm$\tiny{0.0435}} & \makecell{0.0641 \\ \footnotesize$\pm$\tiny{0.0032}} & \makecell{0.0524 \\ \footnotesize$\pm$\tiny{0.0022}}  \\
                        GPF-plus & \makecell{0.3109 \\ \footnotesize$\pm$\tiny{0.1086}} & \makecell{0.2046 \\ \footnotesize$\pm$\tiny{0.0736}} & \makecell{0.5239 \\ \footnotesize$\pm$\tiny{0.0720}} & \makecell{0.3112 \\ \footnotesize$\pm$\tiny{0.0626}}  & \makecell{0.2112 \\ \footnotesize$\pm$\tiny{0.0102}} & \makecell{0.1808 \\ \footnotesize$\pm$\tiny{0.0216}} & \makecell{0.5042 \\ \footnotesize$\pm$\tiny{0.0559}} & \makecell{0.3323 \\ \footnotesize$\pm$\tiny{0.0347}}  & \makecell{0.3606 \\ \footnotesize$\pm$\tiny{0.1570}} & \makecell{\underline{0.2559} \\ \footnotesize$\pm$\tiny{0.0941}}  & \makecell{0.0687 \\ \footnotesize$\pm$\tiny{0.0078}} & \makecell{0.0519 \\ \footnotesize$\pm$\tiny{0.0031}} \\
                        All-in-One & \makecell{0.2337 \\ \footnotesize$\pm$\tiny{0.0878}} & \makecell{0.0982 \\ \footnotesize$\pm$\tiny{0.0418}} & \makecell{0.4310 \\ \footnotesize$\pm$\tiny{0.2792}} & \makecell{0.2317 \\ \footnotesize$\pm$\tiny{0.1106}} & \makecell{0.2116 \\ \footnotesize$\pm$\tiny{0.0217}} & \makecell{0.1902 \\ \footnotesize$\pm$\tiny{0.0203}} & \makecell{0.3127 \\ \footnotesize$\pm$\tiny{0.2199}} & \makecell{0.1616 \\ \footnotesize$\pm$\tiny{0.1077}}  & \makecell{\underline{0.3831} \\ \footnotesize$\pm$\tiny{0.1751}} & \makecell{0.1829\\ \footnotesize$\pm$\tiny{0.0666}} & \makecell{0.0559\\ \footnotesize$\pm$\tiny{0.0251}} & \makecell{0.0152 \\ \footnotesize$\pm$\tiny{0.0057}} \\
                        ProNoG & \makecell{0.3366 \\ \footnotesize$\pm$\tiny{0.0349}} & \makecell{0.2218 \\ \footnotesize$\pm$\tiny{0.0371}} & \makecell{0.2789 \\ \footnotesize$\pm$\tiny{0.0466}} & \makecell{0.2060 \\ \footnotesize$\pm$\tiny{0.0290}} & \makecell{0.2154 \\ \footnotesize$\pm$\tiny{0.0083}} & \makecell{0.2064 \\ \footnotesize$\pm$\tiny{0.0113}} & \makecell{0.2620 \\ \footnotesize$\pm$\tiny{0.1002}} & \makecell{0.2155 \\ \footnotesize$\pm$\tiny{0.0666}} & \makecell{0.2986 \\ \footnotesize$\pm$\tiny{0.1067}} & \makecell{0.1869 \\ \footnotesize$\pm$\tiny{0.0382}} & \makecell{0.0748 \\ \footnotesize$\pm$\tiny{0.0159}} & \makecell{0.0566 \\ \footnotesize$\pm$\tiny{0.0144}} \\
                        \cline{1-13}
                        HS-GPPT & \textbf{\makecell{0.5485 \\ \footnotesize$\pm$\tiny{0.0296}}}  & \textbf{\makecell{0.3676 \\ \footnotesize$\pm$\tiny{0.0173}}} & \textbf{\makecell{0.5887 \\ \footnotesize$\pm$\tiny{0.0423}}} & \textbf{\makecell{0.4428 \\ \footnotesize$\pm$\tiny{0.0585}}}  & \textbf{\makecell{0.2425 \\ \footnotesize$\pm$\tiny{0.0213}}} & \textbf{\makecell{0.2307 \\ \footnotesize$\pm$\tiny{0.0217}}} & \textbf{\makecell{0.6085 \\ \footnotesize$\pm$\tiny{0.1061}}} & \textbf{\makecell{0.5106 \\ \footnotesize$\pm$\tiny{0.0774}}} & \textbf{\makecell{0.5155 \\ \footnotesize$\pm$\tiny{0.1115}}} & \textbf{\makecell{0.4247 \\ \footnotesize$\pm$\tiny{0.0410}}} & \textbf{\makecell{0.3609 \\ \footnotesize$\pm$\tiny{0.0710}}} & \textbf{\makecell{0.3272 \\ \footnotesize$\pm$\tiny{0.0734}}}
                         \\ \bottomrule
 \end{tabular}}
\caption{Accuracy and F1 score under inductive learning.}
\label{tab:appendix_inductive}
\end{table*}

\begin{table*}
\centering
\scalebox{0.8}{
 \begin{tabular}{c|c|c|c|c|c}
 \toprule
 Datasets    & \multicolumn{1}{c|}{Cora} & \multicolumn{1}{c|}{Pubmed} & \multicolumn{1}{c|}{Citeseer} & Ratings & Empire \\ \cline{1-6} \hline \hline            
                        GPPT & GraphMAE & GraphMAE  & EdgePredGPPT  & EdgePredGPPT & EdgePredGPPT \\
                        GPrompt  & GraphMAE & GraphMAE  & EdgePredGPPT  & GraphCL & GraphMAE \\
                        GPF-plus  & GraphMAE & EdgePredGPPT  & GraphMAE  & GraphCL & EdgePredGPPT\\
                        All-in-One  & GraphMAE & GraphMAE  & GraphMAE & GraphCL & GraphCL \\
                        ProNoG  & GraphCL & GraphCL & GraphCL & GraphCL & GraphCL \\
                        \bottomrule
 \end{tabular}}
\caption{Optimal pre-training and prompt tuning combinations on homophilic graphs and large-scale heterophilic graphs under transductive learning. Table entries reflect prompt tuning methods paired with pre-training strategies that achieve the highest performance on each dataset, based on our exhaustive experimental results.}
\label{tab:appendix_combination_trans_1}
\end{table*}

\begin{table*}
\centering
\scalebox{0.8}{
 \begin{tabular}{c|c|c|c|c|c}
 \toprule
 Datasets   & \multicolumn{1}{c|}{Cornell}  & \multicolumn{1}{c|}{Texas} & \multicolumn{1}{c|}{Wisconsin} & \multicolumn{1}{c|}{Chameleon}  & \multicolumn{1}{c}{Squirrel} \\ \cline{1-6} \hline \hline            
                        GPPT  & DGI & GraphCL & SimGRACE & DGI & EdgePredGPPT  \\
                        GPrompt   & GraphCL & DGI & GraphMAE & GraphMAE & EdgePredGPrompt  \\
                        GPF-plus   & EdgePredGPrompt & EdgePredGPrompt & GraphMAE & GraphCL & EdgePredGPrompt \\
                        All-in-One   & EdgePredGPPT & SimGRACE & SimGRACE & GraphCL & GraphCL \\
                        ProNoG  & GraphCL & DGI & GraphCL & GraphCL & DGI \\
                        \bottomrule
 \end{tabular}}
\caption{Optimal pre-training and prompt tuning combinations on heterophilic graphs under transductive learning.}
\label{tab:appendix_combination_trans_2}
\end{table*}

\begin{table*}
\centering
\scalebox{0.8}{
 \begin{tabular}{c|c|c|c|c|c|c}
 \toprule
 Source   & Texas  & Wisconsin & Chameleon & Pubmed   & Squirrel & Ratings \\ 
  Target  & Wisconsin   & Texas  & Squirrel & Texas    & Cornell & Empire \\ \cline{1-7}       
                        GPPT & GraphCL & GraphCL  & DGI  & GraphCL & EdgePredGprompt & EdgePredGPPT \\
                        GPrompt & GraphMAE & SimGRACE  & GraphCL & GraphMAE  & GraphMAE & GraphMAE \\
                        GPF-plus & GraphMAE & GraphCL  & EdgePredGprompt & GraphMAE  & GraphMAE & EdgePredGPPT \\
                        All-in-One & DGI & SimGRACE  & SimGRACE & DGI  & EdgePredGPPT & SimGRACE \\
                        ProNoG  & GraphCL & DGI & DGI & GraphCL & GraphCL & GraphCL \\
                         \bottomrule
 \end{tabular}}
\caption{Optimal pre-training method combinations for each prompt tuning method and dataset under inductive learning.}
\label{tab:appendix_combination_inductive}
\end{table*}

In this section, we present the complete accuracy and F1 score results. We present the accuracy and F1 score results under transductive learning in Table \ref{tab:appendix_homo_trans} and Table \ref{tab:appendix_hete_trans}, and results under inductive learning in Table \ref{tab:appendix_inductive}. These results are consistent with our conclusion, thereby demonstrating the effectiveness of our model. The corresponding pre-training and prompt tuning combinations on each dataset are presented in Table \ref{tab:appendix_combination_trans_1}, Table \ref{tab:appendix_combination_trans_2}, and Table \ref{tab:appendix_combination_inductive}.

\subsection{Results of Baselines with Other Backbones}
\label{section:appendix_graphtransformer}
\begin{table*}
\centering
\scalebox{0.77}{
 \begin{tabular}{c|c|c|c|c|c|c|c|c|c|c}
 \toprule
 Datasets    & \multicolumn{1}{c|}{Cora} & \multicolumn{1}{c|}{Pubmed} & \multicolumn{1}{c|}{Citeseer} & \multicolumn{1}{c|}{Cornell}  & \multicolumn{1}{c|}{Texas} & \multicolumn{1}{c|}{Wisconsin} & \multicolumn{1}{c|}{Chameleon}  & \multicolumn{1}{c|}{Squirrel} & Ratings & Empire \\ \cline{1-11} \hline \hline            
                        DGI & 0.1549 & 0.3622 & 0.1272 & 0.1463 & 0.2111  & 0.1814  & 0.2430 & 0.1797 & 0.1359 & 0.0703 \\
                        GraphCL  & 0.4882 & 0.4506 & 0.3466 & 0.1683 & 0.2300 & 0.2331  & 0.2431 & 0.1934 & \textbf{0.2013} & 0.0774 \\ 
                        SimGRACE & 0.5151 & 0.4528 & 0.2859 & 0.2182 & 0.2023 & 0.1542 & 0.2706 & 0.1959 & 0.1874 & 0.0584 \\
                        \cline{1-11}
                        GPPT & 0.5221 & 0.6012 & 0.3731 & 0.2033 & 0.2702 & 0.1957 & 0.2933 & 0.1960 & 0.1889 & 0.0750 \\
                        GPrompt  & 0.5571 & 0.6442 & 0.3898 & 0.1913 & 0.1874 & 0.1602 & 0.2455 & 0.2064 & 0.1969 & 0.0513 \\
                        GPF-plus  & 0.5571 & 0.3665 & 0.3237 & 0.1830 & 0.1858 & 0.1584 & 0.2340 & 0.1395 & 0.1754 & 0.0396 \\
                        All-in-One  & 0.3830 & 0.4567 & 0.2350 & 0.1456 & 0.2257 & 0.1315 & 0.2208 & 0.1919 & 0.1884 & 0.0364 \\
                        \cline{1-11}
                        HS-GPPT  & \textbf{0.6915} & \textbf{0.6910} & \textbf{0.5043} & \textbf{0.3875} & \textbf{0.5724} & \textbf{0.4554} & \textbf{0.3324} & \textbf{0.2536} & 0.1972 & \textbf{0.3520} \\ \bottomrule
 \end{tabular}}
\caption{The F1 score results of the baseline using UniMP as the backbone under transductive learning.}
\label{tab:appendix_graphtransfomer_trans}
\end{table*}

\begin{table*}
\centering
\scalebox{1}{
 \begin{tabular}{c|c|c|c|c|c|c}
 \toprule
 Source   & Texas  & Wisconsin & Chameleon & Pubmed   & Squirrel & Ratings \\ 
  Target  & Wisconsin   & Texas  & Squirrel & Texas    & Cornell & Empire \\ \cline{1-7}       
                        DGI & 0.1289 &  0.2291  & 0.1876 & 0.1934  & 0.1719 & 0.0299 \\
                        GraphCL & 0.1584 & 0.1828 & 0.1797 & 0.1663  & 0.1706 & 0.0545 \\
                        SimGRACE & 0.2070 & 0.2811  & 0.1996 & 0.2017  & 0.1531 & 0.0730\\
                        \cline{1-7}
                        GPPT & 0.2303 & 0.1798  & 0.1939  & 0.2164  & 0.2084 & 0.0603 \\
                        GPrompt & 0.2060 & 0.2429  & 0.2052 & 0.1952  & 0.1542 & 0.0459\\
                        GPF-plus & 0.1497 & 0.1914  & 0.1692 & 0.1503  & 0.0683 & 0.0369 \\
                        All-in-One & 0.0958 & 0.1269  & 0.1935 & 0.2108  & 0.1661 & 0.0296\\
                        \cline{1-7}
                        HS-GPPT  & \textbf{0.3676} & \textbf{0.4428} & \textbf{0.2307} & \textbf{0.5106} & \textbf{0.4247} & \textbf{0.3272}
                         \\ \bottomrule
 \end{tabular}}
\caption{The F1 score results of the baseline using UniMP as the backbone under inductive learning.}
\label{tab:appendix_graphtransfomer_inductive}
\end{table*}

\begin{table*}
\centering
\scalebox{0.77}{
 \begin{tabular}{c|c|c|c|c|c|c|c|c|c|c}
 \toprule
 Datasets    & \multicolumn{1}{c|}{Cora} & \multicolumn{1}{c|}{Pubmed} & \multicolumn{1}{c|}{Citeseer} & \multicolumn{1}{c|}{Cornell}  & \multicolumn{1}{c|}{Texas} & \multicolumn{1}{c|}{Wisconsin} & \multicolumn{1}{c|}{Chameleon}  & \multicolumn{1}{c|}{Squirrel} & Ratings & Empire \\ \cline{1-11} \hline \hline            
                        DGI  & 0.3126 & 0.5401 & 0.2304 & 0.1474 & 0.1890 & 0.2029 & 0.2647 & 0.1881 & 0.1473 & 0.0873 \\
                        GraphCL & 0.4807 & 0.4872 & 0.3805 & 0.1750 & 0.2151 & 0.2743 & 0.2591 & 0.1908 & 0.1969 & 0.0801 \\ 
                        SimGRACE & 0.4574 & 0.4445 & 0.3276 & 0.1945 & 0.2840 & 0.2499 & 0.2406 & 0.1804 & 0.1832 & 0.0756 \\
                        \cline{1-11}
                        GPPT  & 0.5255 & 0.6491 & 0.3086 & 0.1954 & 0.2345 & 0.1955 & 0.3027 & 0.2134 & 0.1896 & 0.0436 \\
                        GPrompt  & 0.5015 & 0.5966 & 0.4182 & 0.1830 & 0.2273 & 0.1855 & 0.2433 & 0.1873 & \textbf{0.1974} & 0.0678 \\
                        GPF-plus & 0.5544 & 0.6067 & 0.4438 & 0.1773 & 0.2442 & 0.1608 & 0.2568 & 0.1743 & 0.1865 & 0.0676 \\
                        All-in-One & 0.5654 & 0.6443 & 0.2747 & 0.0952 & 0.1613 & 0.1473 & 0.2874 & 0.1943 & 0.1629 & 0.0456 \\
                        \cline{1-11}
                        HS-GPPT  & \textbf{0.6915} & \textbf{0.6910} & \textbf{0.5043} & \textbf{0.3875} & \textbf{0.5724} & \textbf{0.4554} & \textbf{0.3324} & \textbf{0.2536} & 0.1972 & \textbf{0.3520} \\ \bottomrule
 \end{tabular}}
\caption{The F1 score results of the baseline using BWGNN as the backbone under transductive learning.}
\label{tab:appendix_bwgnn_trans}
\end{table*}

\begin{table*}
\centering
\scalebox{1}{
 \begin{tabular}{c|c|c|c|c|c|c}
 \toprule
 Source   & Texas  & Wisconsin & Chameleon & Pubmed   & Squirrel & Ratings \\ 
  Target  & Wisconsin   & Texas  & Squirrel & Texas    & Cornell & Empire \\ \cline{1-7}       
                        DGI  & 0.2143 & 0.1838 & 0.1715 & 0.2043 & 0.1763 & 0.0493\\
                        GraphCL & 0.1524 & 0.1575 & 0.2016 & 0.1968 & 0.1865 & 0.0578 \\
                        SimGRACE & 0.2615 & 0.2485 & 0.1905 & 0.1768 & 0.1685 & 0.0460 \\
                        \cline{1-7}
                        GPPT & 0.1931 & 0.2361 & 0.1639 & 0.2349 & 0.2065 & 0.0278 \\
                        GPrompt & 0.1444 & 0.1871 & 0.1758 & 0.2156 & 0.1658 & 0.0640 \\
                        GPF-plus & 0.1657 & 0.2128 & 0.1746 & 0.2225 & 0.1978 & 0.0450 \\
                        All-in-One & 0.1820 &  0.1743 & 0.1933 & 0.1846 & 0.0963 & 0.0329\\
                        \cline{1-7}
                        HS-GPPT  & \textbf{0.3676} & \textbf{0.4428} & \textbf{0.2307} & \textbf{0.5106} & \textbf{0.4247} & \textbf{0.3272}
                         \\ \bottomrule
 \end{tabular}}
\caption{The F1 score results of the baseline using BWGNN as the backbone under inductive learning.}
\label{tab:appendix_bwgnn_inductive}
\end{table*}

In this section, we present the results of the baselines that use a GraphTransformer model UniMP \cite{shi2020masked} and our used BWGNN \cite{tang2022rethinking} as their backbone. To avoid excessive computational costs from grid searches, we utilize optimal pre-training and prompt-tuning combinations from Tables \ref{tab:appendix_combination_trans_1}, Table \ref{tab:appendix_combination_trans_2}, and Table \ref{tab:appendix_combination_inductive}. The results of transductive learning are shown in Table \ref{tab:appendix_graphtransfomer_trans} and \ref{tab:appendix_bwgnn_trans}, and those of inductive learning are presented in Table \ref{tab:appendix_graphtransfomer_inductive} and \ref{tab:appendix_bwgnn_inductive}. The results indicate that our model outperforms the baselines with various backbones. 

\subsection{Ablation Study}
\label{section:appendix_ablation}

\begin{table*}
\centering
\scalebox{0.74}{
 \begin{tabular}{c|c|c|c|c|c|c|c|c|c|c}
 \toprule
 Datasets    & Cora & Pubmed & Citeseer & Cornell  & Texas & Wisconsin & Chameleon  & Squirrel & Ratings & Empire \\ \cline{1-11} \hline \hline   
 HS-GPPT  & \textbf{0.6915} & \textbf{0.6910} & \textbf{0.5043} & \textbf{0.3875} & \textbf{0.5724} & \textbf{0.4554} & \textbf{0.3324} & \textbf{0.2536} & \textbf{0.1972} & \textbf{0.3520} \\ \cline{1-11}
low-pass & 0.3187 & 0.3743 & 0.2235 & 0.1650 & 0.3031  & 0.2293  & 0.2692 & 0.1989 & 0.1919 & 0.1884 \\ \cline{1-11}
single prompt & 0.6900 & 0.6886 & 0.5036 & 0.3796 & 0.5038 & 0.4355 & 0.3140 & 0.2285 & 0.1960 & 0.3502 \\
\textit{w/o} prompt  & 0.6896 & 0.6893 & 0.4764 & 0.2869 & 0.4123 & 0.3124 & 0.3083 & 0.2255 & 0.1950 & 0.2833 \\
\textit{w/o} prompt norm  & 0.6894 & 0.6873 & 0.4828 & 0.2793 & 0.3260 & 0.4439 & 0.2955 & 0.2484 & 0.1968 & 0.3506 \\ \bottomrule
 \end{tabular}}
\caption{Complete ablation study results under transductive learning.}
\label{tab:appendix_ablation_trans}
\end{table*}

\begin{table*}
\centering
\scalebox{1}{
 \begin{tabular}{c|c|c|c|c|c|c}
 \toprule
 Source   & Texas & Wisconsin & Chameleon & Pubmed   & Squirrel & Ratings  \\ 
 Target  & Wisconsin & Texas & Squirrel & Texas   & Cornell & Empire  \\ \cline{1-7} \hline \hline   
 HS-GPPT  & \textbf{0.3676}  & \textbf{0.4428} & \textbf{0.2307} & \textbf{0.5106} & \textbf{0.4247} & \textbf{0.3272}  \\ \cline{1-7}
low-pass & 0.2811  & 0.3512 & 0.2074 & 0.3323 & 0.2061 & 0.1498   \\ \cline{1-7}
single prompt & 0.3624  & 0.4198 & 0.2204 & 0.5070 & 0.4203 & 0.3271 \\
\textit{w/o} prompt  & 0.2923  & 0.3819 & 0.2077 & 0.4121 & 0.3360 & 0.2518  \\
\textit{w/o} prompt norm  & 0.2820  & 0.2938 & 0.2211 & 0.3011 & 0.3022 & 0.3241 \\ \bottomrule
 \end{tabular}}
\caption{Complete ablation study results under inductive learning.}
\label{tab:appendix_ablation_inductive}
\end{table*}

In this section, we provide the complete results of the ablation study in Table \ref{tab:appendix_ablation_trans} and Table \ref{tab:appendix_ablation_inductive}, which are consistent with our conclusion, demonstrating the effectiveness of each key module.

\end{document}